\documentclass[12pt,letterpaper]{article}
\usepackage[T1]{fontenc}
\usepackage[utf8]{inputenc}
\usepackage{lmodern}
\usepackage[margin=1in]{geometry}
\usepackage{setspace,lineno,microtype}
\usepackage{titlesec}
\usepackage{array,longtable,booktabs,ragged2e,caption,setspace}
\usepackage{threeparttable}
\usepackage{graphicx}
\graphicspath{{figures/}{./}}
\usepackage{amsmath,amssymb}
\usepackage{xcolor}
\usepackage[super,sort&compress,numbers]{natbib}
\usepackage[colorlinks=true,linkcolor=blue!60!black,citecolor=blue!60!black,urlcolor=blue!60!black]{hyperref}
\hypersetup{pdftitle={Instructor-designed generative AI improves learning in university classes}}
\usepackage[normalem]{ulem}

\usepackage{arydshln} % load after array and longtable
\newlength{\surveyUsableWidth}

\newcommand{\surveyThemeShare}{0.17}
\newcommand{\surveyCountShare}{0.09}
\newcommand{\surveyKeywordShare}{0.20}
\newcommand{\surveyQuoteShare}{0.54}


\newcolumntype{L}[1]{>{\RaggedRight\arraybackslash}p{#1}}
\newcolumntype{C}[1]{>{\Centering\arraybackslash}p{#1}}
\renewcommand{\arraystretch}{1.08}

\newcolumntype{P}[1]{>{\RaggedRight\arraybackslash}p{#1}}

\titleformat{\section}{\normalfont\normalsize\bfseries}{\thesection}{0.5em}{}
\titleformat{\subsection}{\normalfont\normalsize\bfseries}{\thesubsection}{0.5em}{}
\titleformat{\subsubsection}{\normalfont\normalsize\bfseries}{\thesubsubsection}{0.5em}{}
\titleformat{\paragraph}[runin]{\normalfont\normalsize\bfseries}{\theparagraph}{0.5em}{}
\titleformat{\subparagraph}[runin]{\normalfont\normalsize\bfseries}{\thesubparagraph}{0.5em}{}
\titlespacing*{\section}{0pt}{9pt}{3pt}
\titlespacing*{\subsection}{0pt}{6pt}{2pt}
\titlespacing*{\subsubsection}{0pt}{6pt}{2pt}
\titlespacing*{\paragraph}{0pt}{4pt}{0.5em}
\titlespacing*{\subparagraph}{0pt}{4pt}{0.5em}
\makeatletter

\renewcommand{\@maketitle}{%
  \begingroup
  \centering
  \setlength{\parskip}{0pt}%
  {\LARGE\@title\par}%
  \vspace{6pt}%
  {\normalsize\@author\par}%
  \vspace{7pt}%
  {\fontsize{10}{11.5}\selectfont\raggedright\paperaffiliations\par}%
  \vspace{4pt}%
  {\fontsize{10}{11.5}\selectfont\raggedright\paperequalcontribution\par}%
  \vspace{4pt}%
  {\fontsize{10}{11.5}\selectfont\raggedright\papercorrespondence\par}%
  \ifx\@date\@empty\else
    \vspace{3pt}{\normalsize\@date\par}%
  \fi
  \endgroup
  \vspace{8pt}%
}
\makeatother

\title{Instructor-designed AI improves learning in a randomized trial across 19 universities}
\author{%
  Jun~Li\textsuperscript{1,\textdagger,*},
  Robert~Bray\textsuperscript{2,\textdagger},  S{\'e}bastien~Martin\textsuperscript{2,\textdagger},  Nils~Van~den~Steen\textsuperscript{3,\textdagger},
  Andrew~Wu\textsuperscript{1,\textdagger},
  Xiao~Ni\textsuperscript{1, 20},
  Arielle~Anderer\textsuperscript{4},
  Ekaterina~Astashkina\textsuperscript{1},
  Volodymyr~O~Babich\textsuperscript{5},
  Jean-Paul~Baldwin\textsuperscript{1},
  Hamsa~Bastani\textsuperscript{6},
  Christian~Blanco\textsuperscript{7},
  Oguzhan~Celebi\textsuperscript{1},
  Christopher~Chen\textsuperscript{8},
  Shi~Chen\textsuperscript{9},
  Benjamin~Collier\textsuperscript{10},
  Yao~Cui\textsuperscript{4},
  Pnina~Feldman\textsuperscript{11},
  Blair~Flicker\textsuperscript{12},
  Brian~Han\textsuperscript{13},
  Yichun~Hu\textsuperscript{4},
  Daniela~Hurtado-Lange\textsuperscript{2},
  Jane~Jiang\textsuperscript{7},
  Zhaohui~Jiang\textsuperscript{10},
  Ismail~Kirci\textsuperscript{13},
  Murat~K{\"o}ksalan\textsuperscript{1},
  Zhen~Lian\textsuperscript{14},
  Hongyan~Liang\textsuperscript{13},
  Mengshi~Lu\textsuperscript{15},
  Lesley~Meng\textsuperscript{14},
  Ken~Moon\textsuperscript{4},
  Mi~Kyong~Newsom\textsuperscript{7},
  Minje~Park\textsuperscript{16},
  Chris~Parker\textsuperscript{11},
  Anyan~Qi\textsuperscript{17},
  Meng~Qi\textsuperscript{4},
  Debjit~Roy\textsuperscript{18},
  Neha~Sharma\textsuperscript{6},
  Alp~Sungu\textsuperscript{6},
  Tom~Tan\textsuperscript{3},
  Necati~Tereyagoglu\textsuperscript{12},
  Hanu~Tyagi\textsuperscript{13},
  Sriram~Venkataraman\textsuperscript{12},
  Guihua~Wang\textsuperscript{17},
  Yixin~Wang\textsuperscript{13},
  Owen~Wu\textsuperscript{8},
  Zonghao~Yang\textsuperscript{19},
  Qiuping~Yu\textsuperscript{5},
  Sophie~Yu\textsuperscript{6},
  Jingwei~Zhang\textsuperscript{4},
  Wenchang~Zhang\textsuperscript{8} \&
  Ruihao~Zhu\textsuperscript{4}
}

\newcommand{\paperaffiliations}{%
  \textsuperscript{1}University of Michigan, Ann Arbor, MI 48109, USA\par
  \textsuperscript{2}Northwestern University, Evanston, IL 60208, USA\par
  \textsuperscript{3}Southern Methodist University, Dallas, TX 75275, USA\par
  \textsuperscript{4}Cornell University, Ithaca, NY 14853, USA\par
  \textsuperscript{5}Georgetown University, Washington, DC 20057, USA\par
  \textsuperscript{6}University of Pennsylvania, Philadelphia, PA 19104, USA\par
  \textsuperscript{7}The Ohio State University, Columbus, OH 43210, USA\par
  \textsuperscript{8}Indiana University Bloomington, Bloomington, IN 47405, USA\par
  \textsuperscript{9}University of Washington, Seattle, WA 98195, USA\par
  \textsuperscript{10}Carnegie Mellon University, Pittsburgh, PA 15213, USA\par
  \textsuperscript{11}University of Virginia, Charlottesville, VA 22903, USA\par
  \textsuperscript{12}University of South Carolina, Columbia, SC 29208, USA\par
  \textsuperscript{13}University of Illinois Urbana--Champaign, Champaign, IL 61820, USA\par
  \textsuperscript{14}Yale University, New Haven, CT 06511, USA\par
  \textsuperscript{15}Purdue University, West Lafayette, IN 47907, USA\par
  \textsuperscript{16}The University of Hong Kong, Hong Kong, China\par
  \textsuperscript{17}The University of Texas at Dallas, Richardson, TX 75080-3021, USA\par
  \textsuperscript{18}Indian Institute of Management Ahmedabad, Ahmedabad, Gujarat 380015, India\par
  \textsuperscript{19}Stevens Institute of Technology, Hoboken, NJ 07030, USA\par
    \textsuperscript{20}Fudan University, Shanghai, China\par
}
\newcommand{\paperequalcontribution}{%
  \textsuperscript{\textdagger}Led by J. Li, the members of this primary research team share first authorship.
}
\newcommand{\papercorrespondence}{%
  \textsuperscript{*}Email: \href{mailto:junwli@umich.edu}{\nolinkurl{junwli@umich.edu}}.
}

\date{}

\begin{document}
%\linenumbers
\maketitle

% =====================================================================
% SUMMARY PARAGRAPH
% =====================================================================
\clearpage
\begin{spacing}{1}

\noindent \textbf{Summary: } Universities are alarmed by recent evidence that artificial intelligence (AI) can undermine learning, with students prescribed generic chatbots or centrally designed AI tools performing worse\cite{Bastani2025,Fan2026,LiuMaryland2026}. In a pre-registered randomized controlled trial across 7,200 students at 19 universities over three semesters, we tested whether AI designed by frontline instructors can improve learning. We encouraged instructors to design AI solutions they believed would best help their students. Drawing on their unique knowledge of their classes and students, the instructors created a wide array of innovations. We show that these instructor-designed AI innovations improved student learning, measured by AI-proof assessments (proctored, in-person examinations). We found no change in student engagement, measured by course activity records, or satisfaction, measured by course evaluations. Our findings demonstrate that AI can improve student learning when its adoption is frontline-driven, with instructors empowered to design and refine AI solutions for their own classrooms.
\end{spacing}

\vspace{15pt}
% =====================================================================
% INTRODUCTION (unheaded, following Nature convention)
% =====================================================================

When left to their own devices, can educators find ways to improve instruction and increase learning with artificial intelligence (AI)? The answer to this question will largely determine the course of higher education in the next few years. Recent studies and university reports paint a bleak picture of ``tectonic disruption,'' \cite{MIT2026,Cornell2023,Brown2026} as students prescribed with off-the-shelf chatbots or centrally-designed AI tools offload critical cognitive processes associated with learning\cite{kosmyna2025your,MIT2026} and perform worse on corresponding tasks and assignments.\cite{Liu2025,Bastani2025} 

However, this picture is likely incomplete, for two reasons. First, methodologically, many existing studies have limited disciplinary coverage (often a single course or subject)\cite{Nie2025, Bastani2025,ThoeniFryer2026}, limited exposure time (often hours or weeks rather than full academic terms)\cite{Bastani2025,Kestin2025,Fan2026}, lack random assignment\cite{Deng2025}, or they assess performance with narrowly defined tasks\cite{Fan2026} or while AI assistance remains available\cite{Deng2025,Weidlich2025}.
Recent reports thus highlight the scarcity of rigorous analyses conducted over longer periods and across diverse disciplines\cite{Pearson2025,Pearson2026}. More importantly, most existing studies focus on one particular type of AI treatment: one-size-fits-all AI tools---either generic AI chatbots or centrally-planned AI agents---which are not amenable to instructor customization, and typically deployed in a top-down fashion\cite{LiuMaryland2026}.

This is not how teaching innovations typically take place in university classrooms, as higher education is intrinsically decentralized: institutions generally avoid mandating uniform rules across classes. The heterogeneous nature of the classroom necessitates giving instructors discretion to choose their teaching style, content, and tools based on their unique understanding of their students. In this regard, AI is likely no different: successful integration of AI will probably require significant input, design effort, and creativity from individual \textit{frontline instructors}\cite{Kessler2018}. 

This instructor-led integration makes our study unique. Rather than prescribe a single AI tool and mandate a single method for using it, we encourage the instructors in our study to use their knowledge of their students and subjects to design an AI-powered solution that best fits their class; we then provide technological support and a knowledge sharing platform to help them realize and deploy their designs. This approach gives instructors ownership and autonomy over the AI in their classrooms\cite{MoonKessler2026}---they can create something they are proud of, and dynamically refine it in response to student feedback. The instructors in our study developed a wide array of innovative solutions.

\begin{figure}[!hp]
\centering
\includegraphics[width=\linewidth,height=0.74\textheight,keepaspectratio]{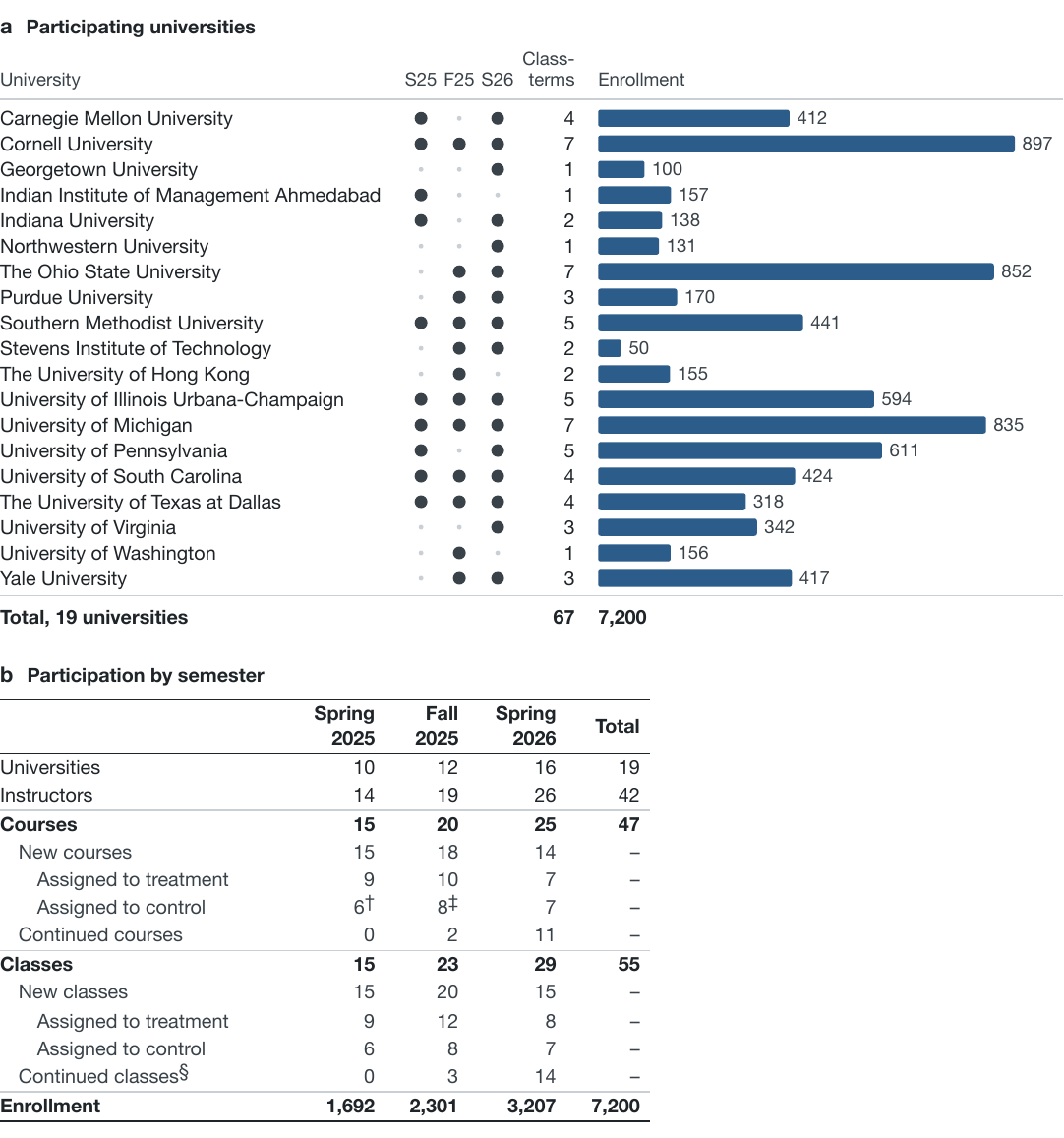}
\caption{\textbf{RCT participation.} A class is a course--instructor pair; a class-term is one term of a class. In \textbf{a}, dots mark the semesters in which a university took part (S25, spring 2025; F25, fall 2025; S26, spring 2026) and bars show enrollment summed over class-terms. In \textbf{b}, totals count by university, instructor, course and class across semesters; enrollment sums semesters. Treatment and control counts refer to new courses or classes. \textsuperscript{\textdagger}Eight courses were assigned to control in Spring 2025; two classes taught by the same instructor withdrew before receiving the assignment. \textsuperscript{\textdaggerdbl}Nine courses were assigned to control in Fall 2015; one class withdrew before receiving its assignment. \textsuperscript{\S}Includes five new course--instructor pairs: the course continued from an earlier term, but with instructors who had not taken part in that term's experiment. Therefore, total classes (55) exceed new classes (50).}
\label{fig-flow}
\end{figure}

Another distinguishing feature of our study is its scope. We conduct pre-registered randomized evaluations across 7,200 students enrolled in 55 applied science, technology, engineering, and mathematics (STEM) classes (i.e., course-instructor pairs) at 19 institutions, over three academic semesters from spring 2025 to spring 2026 (Fig.~\ref{fig-flow}a,b). 
We measure learning, engagement, and satisfaction over full academic terms. We compare student performance with AI-free assessments (for instance, proctored paper-based exams) in the years leading up to and following our experimental intervention, with a difference-in-differences evaluation design. Our study thus responds to the emerging call for a comprehensive evaluation of AI in higher education\cite{Pearson2025,Pearson2026,PasqualeEtAl2025}. 

The participating courses covered a broad range of applied STEM subjects, including statistics, probability, machine learning, optimization, simulation, data analytics and visualization, regression, forecasting, decision analysis, and modeling, with applications in economics, management, health care, supply chain and logistics, information technology, and finance. We grouped classes by subject and academic level (undergraduate or graduate) and randomly assigned classes within groups to implement their instructor-designed AI innovations either immediately (treatment) or with a one-teaching-term delay (control). Following our pre-registered protocol, we excluded classes for which the instructor, curriculum, or assessments meaningfully changed during the study period. For analyses of learning outcomes, we additionally excluded classes whose assessments permitted AI assistance during examinations. The resulting class-level sample comprises 44, 54, and 47 experimental class-terms for learning, engagement, and satisfaction, respectively, whereas the student-level sample comprises 44, 50, and 47. Both samples incorporate an additional 67 pre-experimental class-terms, for historical comparison.

% =====================================================================
% RESULTS
% =====================================================================
\section*{Results}

\subsection*{Instructors found diverse uses for AI} 

With technical support from the research team, the instructors in our study designed and implemented many different AI use cases, tailored to the particulars of their respective classes. These AI interventions---several of which were the first meaningful implementation of their kind in higher education---spanned 33 types, and seven pedagogical functions (Fig.~\ref{fig-interventions}). For example, instructors developed interactive simulations (8 classes) and stakeholder role-play sessions (8 classes), in which students interact with---and actively soliciting information from---AI-simulated personas or organizations. Instead of generic teaching assistant bots, instructors designed individualized coaching agents (19 classes) that used various forms of Socratic questioning, as well as tutors (18 classes) that supported targeted practice, examination preparation, and formative feedback. Many instructors (25 classes) also developed AI-powered course navigation tools that complement traditional learning management systems with personalized material delivery and logistics management. Instructors also designed AI agents to help students outside of the core course content areas, by, for example, having students reflect on the personal relevance of the course (6 classes), or help students develop AI literacy (5 classes). Instructors typically combined a mixture of these use cases, with the average class's AI innovation serving 3.0 distinct pedagogical functions. 

\subsection*{Instructor-designed AI solutions improved learning}
\begin{figure}[p]
\centering
\includegraphics[width=\linewidth]{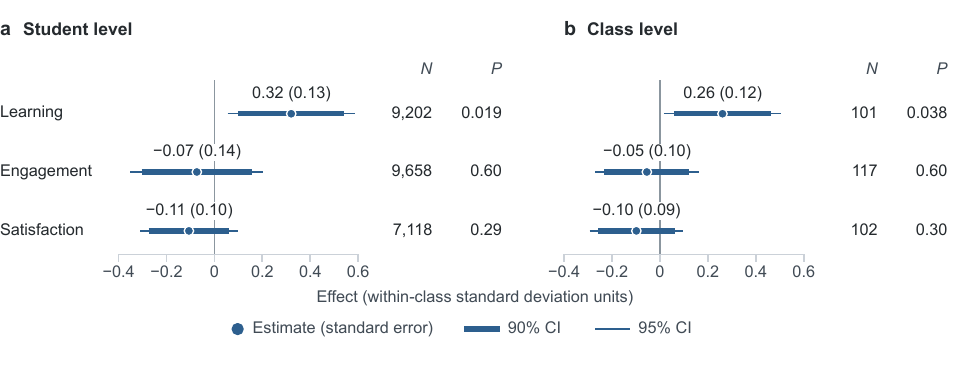}
\caption{\textbf{Effects of instructor-designed AI innovations on learning, engagement, and satisfaction.}  \textbf{a}, Student-level analysis; \textbf{b}, class-level analysis. Points are OLS estimates of the contemporaneous AI-innovation effect ($\beta_0$) from the primary specification (Methods), measured in
within-class standard deviation units, with CR2 standard errors
clustered by class-term in parentheses. Thick bars show 90\% and thin lines 95\% confidence intervals. $N$ is the number of student observations in \textbf{a} and of class-terms, including historical terms, in \textbf{b}. }
\label{fig-main}
\end{figure}

We measured learning with uncurved scores from AI-proof assessments, typically from proctored, in-person, pen-and-paper examinations.  
We standardized learning scores to account for heterogeneity at the class level and to measure the treatment effects in standard deviations. We used a pre-registered difference-in-differences regression to compare changes from historical outcomes between treatment and control classes, accounting for class differences and common time effects. The estimates suggest that the instructor-designed AI innovations significantly improved student learning (Fig.~\ref{fig-main}). The student-level treatment effect estimate is 0.32 standard deviations, SD (SE=0.13; P=0.019; 95\% CI=$[0.06, 0.59]$). The class-level estimate is 0.26 SD (SE=0.12; P=0.038; 95\% CI=$[0.02, 0.51]$).\footnote{
    These estimates stem from regressions that is slightly simpler than the ones we pre-registered (see Methods, Empirical Specifications); however, our pre-registered specifications yield nearly identical results (see Extended Data Fig~\ref{edfig-twelve}).
}  
The results are similar across three estimators (the ordinary least squares (OLS) estimator and two instrumental variable (IV) estimators) and two inference methods (Student's $t$ intervals and CR2-adjusted intervals), at both the student and class levels (Extended Data Fig.~\ref{edfig-twelve}).

\subsection*{No detectable effect on engagement and satisfaction measures}

For student satisfaction measures, we use the numerical approval scores of either the course or instructor provided in the institution's end-of-the-term course evaluations. For student engagement measures, we used the best available data for each class: some instructors tracked attendance or in-class participation, while others tracked homework completion, time spent, or activity duration on their respective learning management systems (such as Canvas or Blackboard). We standardized these heterogeneous measures at the class level. 

We found no statistically significant effects for student engagement or satisfaction (Fig.~\ref{fig-main}). The engagement effect estimates are $-0.07$ SD at the student level (SE=0.14; P=0.60; 95\% CI=$[-0.35, 0.20]$) and $-0.05$ SD at the class level (SE=0.10; P=0.60; 95\% CI=$[-0.27, 0.16]$). The satisfaction effect estimates are $-0.11$ SD at the student level (SE=0.10; P=0.29; 95\% CI=$[-0.31, 0.10]$) and $-0.10$ SD at the class level (SE=0.09; P=0.30; 95\% CI=$[-0.29, 0.10]$). These results are consistent across alternative specifications (Extended Data Fig.~\ref{edfig-twelve}).

\subsection*{Student and instructor perspectives}

\vspace{0.4em}

\noindent \textbf{Student perspectives:} We invited students in all treatment classes to complete an anonymous end-of-course survey, which included both quantitative and open-ended questions. 2,019 out of 4,898 students responded. The students responded to the statements ``I put a lot of effort into this course,'' ``I learned a lot in this course,'' and ``I am likely to recommend this class to a peer'' on a five-point Likert scale, ranging from ``strongly disagree'' to ``strongly agree''. The mean ratings were 4.00, 3.97, and 3.72, respectively.  

The students also rated their AI experience. Overall, the students overwhelmingly rated their experience as positive, with 16.1 times as many students reporting that ``The AI learning aids improved the educational experience'' as reporting that ``The AI learning aids worsened the educational experience,'' and 5.1 times as many reporting that ``Additional AI learning aids would have improved the educational experience'' as reporting that ``Additional AI learning aids would have worsened the educational experience.''

\begin{figure}[p]
\centering
\includegraphics[width=\linewidth]{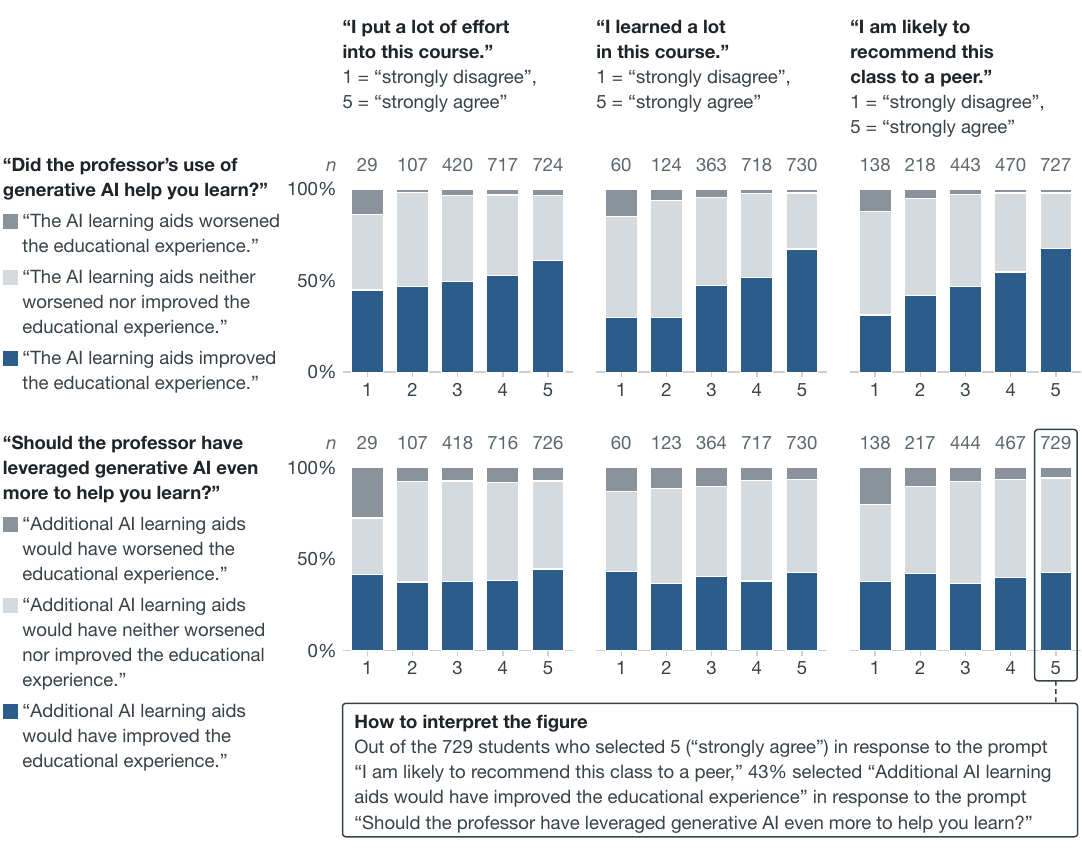}
\caption{\textbf{Student Survey Responses - Structured Questions} Results are based on an anonymous end-of-course survey: 2,019 out of 4,898 students enrolled in treatment classes responded.  }
\label{fig-survey-choice}
\end{figure}

Further, the students' appreciation of the AI innovations correlated strongly with their appreciation of the class (Fig~\ref{fig-survey-choice}). For example, the respective mean scores for prompts ``I learned a lot in this course'' and ``I am likely to recommend this class to a peer'' are 4.18 and 3.98 for students who reported that ``The AI learning aids improved the educational experience,'' but are only 3.35 and 2.96 for students who reported that ``The AI learning aids worsened the educational experience.'' Both of these differences are statistically significant at the $p = 0.0001$ level.

In responses to the open-ended question asking what aspects of the AI innovations they most valued (N=1,414), students most frequently highlighted practice and examination preparation (49.2\%)---they liked having access to ``endless problems,'' and they appreciated the AI's support for recall, comprehension and problem solving (33.0\%), describing interactions that helped ``break down complex topics into simpler explanations.'' Students also mentioned the value of content tailored to their courses (23.2\%): one described tools ``built around the class content---it increased my trust.'' They valued time savings (14.5\%), with one noting, ``I didn't attend office hours for every single minuscule question,'' and interactivity and engagement (6.4\%), describing the experience as ``engaging a different part of my brain.''  Students have also noted that these interactions increased their own AI literacy (2.8\%), including learning ``how different prompts impact outcomes.'' Finally, students  specifically mentioned support for critical thinking (2.0\%): ``I like how this AI learning tool doesn't give answers to my hw questions right away. It gives hints along the way and helped me get to the answer.'' Fig~\ref{fig-survey}a provides the summary and Extended Data Table~\ref{tab:survey-v3-1} provides the detailed quotes.

In response to the open-ended question about what they did not like and how to improve (N=1,157), students most frequently requested more regular AI use throughout the course (23.5\%), including for regular checkpoints ``after every unit,'' and more challenging and varied practice questions (22.7\%). They also sought closer alignment with course content (20.1\%), and clearer, deeper explanations (19.9\%). They suggested ways to improve AI response accuracy (10.5\%), requested more guidance on AI use (11.1\%), such as ``examples of effective prompts,'' and asked for more engaging activities (6.3\%), such as ``more AI simulations.'' Some students (8.6\%) preferred less AI use  in class. Students also expressed concerns about overreliance (4.0\%), and requested continued human support (3.0\%).

\vspace{0.8em}

\noindent \textbf{Instructor perspectives:} We interviewed all 25 instructors who implemented AI in our study. We asked each interviewee (1) what motivated their AI designs, (2) why they considered these designs suitable for their instructional goals, and (3) what they experienced and learned during implementation. 

\begin{figure}[p]
\centering
\includegraphics[width=\linewidth]{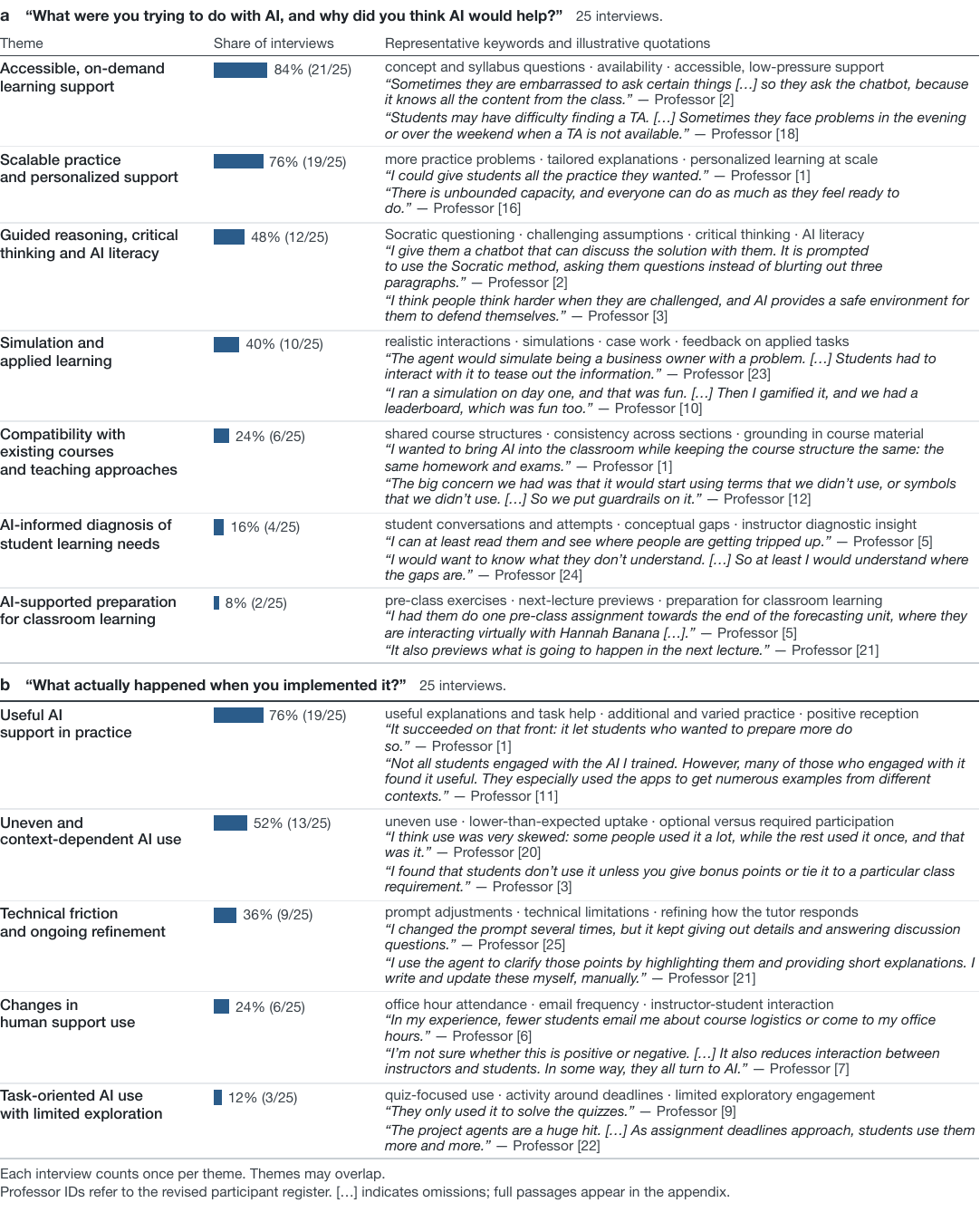}
\caption{\textbf{Instructor interviews: themes.} Share of the 25 instructor interviews featuring each theme, with representative keywords and illustrative quotations. \textbf{a}, What instructors were trying to do with AI, and why they thought AI would help. \textbf{b}, What happened when they implemented it. Each interview counts once per theme; themes may overlap.}
\label{fig-faculty}
\end{figure}

Qualitative analysis of the instructors' answers to questions (1) and (2) identified seven themes concerning their objectives for using AI and its perceived benefits. Instructors highlighted accessible, on-demand learning support (84\%), including help ``in the evening or over the weekend when a TA is not available'' (Professor 18). They also emphasized scalable practice and personalized support (76\%): ``Here, you have unbounded capacity, and everybody can do as much as they feel ready to do'' (Professor 16). For guided reasoning, critical thinking, and AI literacy (48\%), one instructor ``prompted [a chatbot] to use the Socratic method, asking them questions instead of blurting out three paragraphs'' (Professor 2). Instructors also emphasized using AI to deliver more interactive simulation and applied learning (40\%) in class, including gamified activities: ``We had a leaderboard and a very gamified version of the DuPont model'' (Professor 10). Instructors also sought compatibility with existing courses and teaching approaches (24\%), aiming to ``bring AI into the classroom while keeping the structure of the course the same'' (Professor 1). They also reported using the AI chat logs to identify student learning difficulties (16\%), to ``see where people are getting tripped up'' (Professor 5), and they used AI activities to prepare students for classroom learning (8\%), for example by previewing ``what is going to happen in the next lecture'' (Professor 21). Fig.~\ref{fig-faculty}a summarizes these overlapping themes, and Extended Data Table~\ref{tab:faculty-intentions} provides detailed accounts.

Qualitative analysis of the instructors' answers to question (3) identified five themes concerning actual student use, surprises, and adjustments during implementation. The majority of instructors reported positive experiences with AI support (76\%): ``It succeeded on that front: it let students who wanted to prepare more do so'' (Professor 1). They also described uneven and context-dependent AI use (52\%), with one instructor observing that ``some people used it a lot, while the rest used it once'' (Professor 20). Instructors described technical friction and ongoing refinement (36\%), including difficulties controlling AI responses and the need to fine-tune prompts: ``I changed the prompt several times, but it kept giving out details and answering discussion questions'' (Professor 25). They also noted changes in students' use of human support (24\%), with one reporting ``fewer students email me about course logistics or come to my office hours'' (Professor 6). Finally, instructors reported task-oriented AI use (12\%), with one observing increased use typically around deadlines: ``As assignment deadlines approach, students use them more and more'' (Professor 22; Fig.~\ref{fig-faculty}b; Extended Data Table~\ref{tab:faculty-implementation}). 

% =====================================================================
\section*{Discussion}
% =====================================================================

Universities have recently sounded the alarm that AI might undermine student learning and cognitive development \cite{MIT2026,Brown2026,Cornell2023}. However, after adoption of the instructor-led AI in our multi-site randomized trial, student learning outcomes \emph{increased} by 0.26 SD at the class level and by 0.32 SD at the student level. These effects are sizable: for comparison, a study of pre-K--12 randomized interventions reported a median improvement of 0.10 SD\cite{Kraft2020}, and an analysis of active learning in undergraduate STEM reported a mean improvement of 0.47 SD\cite{Freeman2014}.

Our positive learning estimates contradict the negative estimates found in the other major studies on AI in education\cite{Bastani2025,Fan2026,LiuMaryland2026}. Our study's key differentiator is its use of frontline-driven AI. Whereas other studies use a top-down approach, imposing a uniform protocol devised by the researchers, we use a bottom-up approach, spurring instructors to develop AI solutions to problems faced in their particular classes. Empowering instructors to create and continuously improve their own AI innovations leverages their unique knowledge of their classes and their students\cite{KepplerLiWu2022}.

The AI innovations addressed longstanding challenges in higher education. Large enrollments limit opportunities for individualized dialogue and feedback\cite{Nicol2010Dialogue}, and scheduling conflicts and social pressure further   discourage students from seeking academic help\cite{Hsu2022OfficeHours,Karabenick2003LargeClasses}. The instructors in our study sought to overcome these constraints by using AI to provide personalized explanations, guided practice, and feedback beyond scheduled class hours. Timely, actionable feedback may help students identify gaps, adjust their approach, and continue practicing while they still have the opportunity and inclination to improve\cite{NicolMacfarlaneDick2006Feedback,Shute2008Feedback}. Furthermore, requiring students to attempt problems and explain their reasoning before receiving answers, as most of the instructors in our study chose to do, may help preserve the constructive engagement central to learning\cite{ChiWylie2014ICAP}. Instructors also used AI to support active and experiential learning through activities such as real-time stakeholder roleplay, allowing students to apply concepts, make decisions, and reflect on their consequences \cite{Freeman2014,Burch2019Experiential,BenZion2026PhysicsSimulations}. Finally, as surfaced from our surveys and interviews, AI implementation that focuses on frontline instructors can also support continuous instructional improvement: students' questions and difficulties can reveal where explanations, prompts, or activities need revision\cite{WiseJung2019Analytics}. 

The results of our study should be interpreted with its limitations in mind. First, the participating classes were primarily applied STEM, and effects in other disciplines and educational settings remain to be tested. Second, few classes were observed during a second or third implementation, limiting conclusions about the persistence of learning gains across repeated course offerings. Third, our engagement measures varied across sites and courses, due to  heterogeneity in data availability and learning management systems. This reflects a longstanding measurement challenge in education research\cite{Fredricks2004}: attendance and recorded participation capture only the in-class dimensions of engagement\cite{WangCasePrep2026}, while learning activity outside the classroom are typically unobserved. Finally, our RCT does not have enough statistical power to distinguish effects across different AI innovations and instructional settings. Future studies should determine which innovations are more effective and in which settings, and identify design features that contribute to targeted learning gains. 
% =====================================================================
% MAIN REFERENCES (Nature order: main text, main references, then Methods)
% =====================================================================
\bibliographystyle{unsrtnat}
\bibliography{references}

@article{BellMcCaffrey2002,
  author  = {Bell, Robert M. and McCaffrey, Daniel F.},
  title   = {Bias Reduction in Standard Errors for Linear Regression with Multi-Stage Samples},
  journal = {Survey Methodology},
  year    = {2002},
  volume  = {28},
  number  = {2},
  pages   = {169--181}
}

@article{ImbensKolesar2016,
  author  = {Imbens, Guido W. and Kolesár, Michal},
  title   = {Robust Standard Errors in Small Samples: Some Practical Advice},
  journal = {The Review of Economics and Statistics},
  year    = {2016},
  volume  = {98},
  number  = {4},
  pages   = {701--712},
  doi     = {10.1162/REST_a_00552}
}

@article{Bray2025,
  author  = {Bray, Robert L.},
  title   = {A Tutorial on Teaching Data Analytics with Generative {AI}},
  journal = {INFORMS Journal on Applied Analytics},
  year    = {2025},
  volume  = {55},
  number  = {4},
  pages   = {319--343},
  doi     = {10.1287/inte.2023.0053}
}

@article{Fan2026,
  author  = {Fan, Yizhou and Tang, Luzhen and Le, Huixiao and Shen, Kejie and Tan, Shufang and Zhao, Yueying and Shen, Yuan and Li, Xinyu and Ga{\v{s}}evi{\'c}, Dragan},
  title   = {Beware of Metacognitive Laziness: Effects of Generative Artificial Intelligence on Learning Motivation, Processes, and Performance},
  journal = {British Journal of Educational Technology},
  year    = {2025},
  volume  = {56},
  number  = {2},
  pages   = {489--530},
  doi     = {10.1111/bjet.13544}
}

@article{Bastani2025,
author = {Bastani, Hamsa and Bastani, Osbert and Sungu, Alp and Ge, Haosen and Kabakcı, {\"{O}}zge and Mariman, Rei},
doi = {10.1073/pnas.2422633122},
issn = {10916490},
journal = {Proceedings of the National Academy of Sciences of the United States of America},
number = {26},
pages = {1--8},
pmid = {40560616},
title = {{Generative AI without guardrails can harm learning: Evidence from high school mathematics}},
volume = {122},
year = {2025}
}

@article{Kestin2025,
author = {Kestin, Greg and Miller, Kelly and Klales, Anna and Milbourne, Timothy and Ponti, Gregorio},
doi = {10.1038/s41598-025-97652-6},
issn = {20452322},
journal = {Scientific Reports},
number = {1},
pages = {17458},
pmid = {40537565},
title = {{AI tutoring outperforms in-class active learning: an RCT introducing a novel research-based design in an authentic educational setting}},
volume = {15},
year = {2025}
}

@article{BenZion2026PhysicsSimulations,
  author    = {Ben-Zion, Yossi and Carroll, Turhan K. and West, Colin G.
               and Wong, Jesse and Finkelstein, Noah D.},
  title     = {Leveraging generative artificial intelligence for
               simulation-based physics experiments: A new approach
               to virtual learning about the real world},
  journal   = {Physical Review Physics Education Research},
  year      = {2026},
  volume    = {22},
  number    = {1},
  pages     = {010109},
  month     = jan,
  publisher = {American Physical Society},
  doi       = {10.1103/s8dy-kqy5},
  url       = {https://doi.org/10.1103/s8dy-kqy5}
}

@article{Deng2025,
author = {Deng, Ruiqi and Jiang, Maoli and Yu, Xinlu and Lu, Yuyan and Liu, Shasha},
doi = {10.1016/j.compedu.2024.105224},
issn = {03601315},
journal = {Computers and Education},
number = {August 2024},
pages = {105224},
publisher = {Elsevier Ltd},
title = {{Does ChatGPT enhance student learning? A systematic review and meta-analysis of experimental studies}},
url = {https://doi.org/10.1016/j.compedu.2024.105224},
volume = {227},
year = {2025}
}

@inproceedings{Nie2025,
  author    = {Nie, Allen and Chandak, Yash and Suzara, Miroslav and Malik, Ali and Woodrow, Juliette and Peng, Matt and Sahami, Mehran and Brunskill, Emma and Piech, Chris},
  title     = {The {GPT} Surprise: Offering Large Language Model Chat in a Massive Coding Class Reduced Engagement But May Increase Adopters' Exam Performances},
  booktitle = {Proceedings of the Twelfth {ACM} Conference on Learning @ Scale (L@S '25)},
  year      = {2025},
  pages     = {376--380},
  publisher = {ACM},
  address   = {New York},
  doi       = {10.1145/3698205.3733960}
}

@article{Liu2025,
  author  = {Liu, Zhiwei and Zuo, Haode and Lu, Yongjing},
  title   = {The Impact of {ChatGPT} on Students' Academic Achievement: A Meta-Analysis},
  journal = {Journal of Computer Assisted Learning},
  year    = {2025},
  volume  = {41},
  number  = {4},
  pages   = {e70096},
  doi     = {10.1111/jcal.70096}
}

@article{Weidlich2025,
  author  = {Weidlich, Joshua and Ga{\v{s}}evi{\'c}, Dragan and Drachsler, Hendrik and Kirschner, Paul A.},
  title   = {{ChatGPT} in Education: An Effect in Search of a Cause},
  journal = {Journal of Computer Assisted Learning},
  year    = {2025},
  volume  = {41},
  number  = {5},
  pages   = {e70105},
  doi     = {10.1111/jcal.70105}
}

@article{Kraft2020,
  author  = {Kraft, Matthew A.},
  title   = {Interpreting Effect Sizes of Education Interventions},
  journal = {Educational Researcher},
  year    = {2020},
  volume  = {49},
  number  = {4},
  pages   = {241--253},
  doi     = {10.3102/0013189X20912798}
}

@techreport{LiuMaryland2026,
  author      = {Liu, Jing and Sweet, Tracy and Chen, Marty Haoyuan and Engelberg, Jonathan and Masters, Megan and Clark, Martyn and Persaud, Axel and Lancaster, Alia and Hollingsworth, Jeffrey and King Rice, Jennifer},
  title       = {The Effects of Course-Integrated {AI} Tutoring on Student Performance and Engagement: A Randomized University Trial},
  institution = {Annenberg Institute at Brown University},
  type        = {EdWorkingPaper},
  number      = {26-1598},
  year        = {2026},
  doi         = {10.26300/y3f8-vh05}
}

@techreport{MIT2026,
  author      = {{Ad Hoc Committee on AI Use in Teaching, Learning, and Research Training}},
  title       = {Report of {MIT}'s Ad Hoc Committee on {AI} Use in Teaching, Learning, and Research Training},
  institution = {Massachusetts Institute of Technology},
  year        = {2026},
  month       = aug,
  url         = {https://aiandeducation.mit.edu/},
  note        = {Co-chairs E. Klopfer and S. Madden. Dated 13 August 2026; released 25 August 2026. Accessed 6 October 2026}
}

@techreport{Cornell2023,
  author      = {Bala, Kavita and Colvin, Alex and others},
  title       = {Generative Artificial Intelligence for Education and Pedagogy},
  institution = {Cornell University},
  type        = {Committee report},
  year        = {2023},
  month       = jul,
  url         = {https://teaching.cornell.edu/generative-artificial-intelligence/cu-committee-report-generative-artificial-intelligence-education},
  note        = {Dated 18 July 2023. Accessed 6 October 2026}
}

@techreport{Brown2026,
  author      = {{Generative AI in Teaching and Learning (GAITL) Committee}},
  title       = {Generative {AI} in Teaching and Learning ({GAITL}) Committee Final Report and Recommendations},
  institution = {Brown University},
  year        = {2026},
  month       = jul,
  url         = {https://provost.brown.edu/sites/default/files/GAITL_Committee_Report_FNL.pdf},
  note        = {Co-chairs E. Kaldor and M. L. Littman. Provost's letter dated 7 July 2026. Accessed 6 October 2026}
}

@article{Freeman2014,
  author  = {Freeman, Scott and Eddy, Sarah L. and McDonough, Miles and Smith, Michelle K. and Okoroafor, Nnadozie and Jordt, Hannah and Wenderoth, Mary Pat},
  title   = {Active learning increases student performance in science, engineering, and mathematics},
  journal = {Proceedings of the National Academy of Sciences},
  year    = {2014},
  volume  = {111},
  number  = {23},
  pages   = {8410--8415},
  doi     = {10.1073/pnas.1319030111}
}

@article{Fredricks2004,
  author  = {Fredricks, Jennifer A. and Blumenfeld, Phyllis C. and Paris, Alison H.},
  title   = {School Engagement: Potential of the Concept, State of the Evidence},
  journal = {Review of Educational Research},
  year    = {2004},
  volume  = {74},
  number  = {1},
  pages   = {59--109},
  doi     = {10.3102/00346543074001059}
}

@article{Pearson2025,
  author  = {Pearson, Helen},
  title   = {Universities are embracing {AI}: will students get smarter or stop thinking?},
  journal = {Nature},
  year    = {2025},
  volume  = {646},
  number  = {8086},
  pages   = {788--791},
  doi     = {10.1038/d41586-025-03340-w},
  note    = {News Feature}
}

@article{Nicol2010Dialogue,
  author = {Nicol, D.},
  year = {2010},
  title = {From monologue to dialogue: improving written feedback processes in mass higher education},
  journal = {Assessment \& Evaluation in Higher Education},
  volume = {35},
  number = {5},
  pages = {501--517},
  doi = {10.1080/02602931003786559},
  url = {https://doi.org/10.1080/02602931003786559},
  citationcount = {1061},
  citationprovider = {OpenAlex},
  citationretrieved = {2026-10-23},
  citationurl = {https://api.openalex.org/works/https://doi.org/10.1080/02602931003786559?select=id,doi,title,cited_by_count,counts_by_year,authorships,primary_location,updated_date},
  journalimpactfactor = {6.9},
  jifyear = {2025},
  jifsource = {https://www.tandfonline.com/action/journalInformation?journalCode=caeh20},
  jifverification = {Publisher or professional association},
  journalstanding = {Established specialist journal for assessment, evaluation and feedback in higher education.},
  authorinstitutions = {University of Strathclyde (UK)},
  affiliationsource = {https://www.tandfonline.com/doi/abs/10.1080/02602931003786559},
}

@article{ChiWylie2014ICAP,
  author = {Chi, M. T. H. and Wylie, R.},
  year = {2014},
  title = {The {ICAP} Framework: Linking Cognitive Engagement to Active Learning Outcomes},
  journal = {Educational Psychologist},
  volume = {49},
  number = {4},
  pages = {219--243},
  doi = {10.1080/00461520.2014.965823},
  url = {https://doi.org/10.1080/00461520.2014.965823},
  citationcount = {2878},
  citationprovider = {OpenAlex},
  citationretrieved = {2026-10-23},
  citationurl = {https://api.openalex.org/works/https://doi.org/10.1080/00461520.2014.965823?select=id,doi,title,cited_by_count,counts_by_year,authorships,primary_location,updated_date},
  journalimpactfactor = {10.6},
  jifyear = {2025},
  jifsource = {https://www.tandfonline.com/action/journalInformation?journalCode=hedp20},
  jifverification = {Publisher or professional association},
  journalstanding = {Leading theory and review journal in educational psychology; published for APA Division15.},
  authorinstitutions = {Arizona State University (USA)},
  affiliationsource = {https://education.asu.edu/sites/default/files/lcl/chiwylie2014icap_2.pdf},
}

@article{Karabenick2003LargeClasses,
  author = {Karabenick, S. A.},
  year = {2003},
  title = {Seeking help in large college classes: A person-centered approach},
  journal = {Contemporary Educational Psychology},
  volume = {28},
  number = {1},
  pages = {37--58},
  doi = {10.1016/S0361-476X(02)00012-7},
  url = {https://doi.org/10.1016/S0361-476X(02)00012-7},
  citationcount = {396},
  citationprovider = {OpenAlex},
  citationretrieved = {2026-10-23},
  citationurl = {https://api.openalex.org/works/https://doi.org/10.1016/S0361-476X\%2802\%2900012-7?select=id,doi,title,cited_by_count,counts_by_year,authorships,primary_location,updated_date},
  journalimpactfactor = {4.4},
  jifyear = {2025},
  jifsource = {https://hub.hku.hk/handle/10722/244730},
  jifverification = {Institutional JCR metadata},
  journalstanding = {Established major research journal in educational psychology.},
  authorinstitutions = {Eastern Michigan University, USA},
  affiliationsource = {https://openalex.org/W2032350718 ; https://files.eric.ed.gov/fulltext/ED454760.pdf ; https://www.sciencedirect.com/science/article/pii/S0361476X02000127},
}

@article{Hsu2022OfficeHours,
  author = {Hsu, J. L. and Rowland-Goldsmith, M. and Schwartz, E. B.},
  year = {2022},
  title = {Student Motivations and Barriers toward Online and In-Person Office Hours in {STEM} Courses},
  journal = {CBE---Life Sciences Education},
  volume = {21},
  number = {4},
  pages = {ar68},
  doi = {10.1187/cbe.22-03-0048},
  url = {https://doi.org/10.1187/cbe.22-03-0048},
  citationcount = {34},
  citationprovider = {OpenAlex},
  citationretrieved = {2026-10-23},
  citationurl = {https://api.openalex.org/works/https://doi.org/10.1187/cbe.22-03-0048?select=id,doi,title,cited_by_count,counts_by_year,authorships,primary_location,updated_date},
  journalimpactfactor = {3.6},
  jifyear = {2025},
  jifsource = {https://www.bioxbio.com/journal/CBE-LIFE-SCI-EDUC},
  jifverification = {Secondary-reported; primary confirmation unavailable},
  journalstanding = {Major specialist venue for discipline-based life-science education research and practice, published by ASCB.},
  authorinstitutions = {Chapman University, USA (all three authors)},
  affiliationsource = {https://pmc.ncbi.nlm.nih.gov/articles/PMC9727601/ ; https://pubmed.ncbi.nlm.nih.gov/36125927/ ; https://doi.org/10.1187/cbe.22-03-0048},
}

@article{NicolMacfarlaneDick2006Feedback,
  author = {Nicol, D. J. and Macfarlane-Dick, D.},
  year = {2006},
  title = {Formative assessment and self-regulated learning: a model and seven principles of good feedback practice},
  journal = {Studies in Higher Education},
  volume = {31},
  number = {2},
  pages = {199--218},
  doi = {10.1080/03075070600572090},
  url = {https://doi.org/10.1080/03075070600572090},
  citationcount = {5808},
  citationprovider = {OpenAlex},
  citationretrieved = {2026-10-23},
  citationurl = {https://api.openalex.org/works/https://doi.org/10.1080/03075070600572090?select=id,doi,title,cited_by_count,counts_by_year,authorships,primary_location,updated_date},
  journalimpactfactor = {4.9},
  jifyear = {2025},
  jifsource = {https://www.tandfonline.com/action/journalInformation?journalCode=cshe20\&show=aimsScope},
  jifverification = {Publisher or professional association},
  journalstanding = {Established major international higher-education research journal, associated with SRHE.},
  authorinstitutions = {University of Strathclyde, UK (Nicol); University of Glasgow, UK (Macfarlane-Dick)},
  affiliationsource = {https://www.tandfonline.com/doi/full/10.1080/03075070600572090 ; https://www.reap.ac.uk/reap/public/Papers/DN_SHE_Final.pdf},
}

@article{Shute2008Feedback,
  author = {Shute, V. J.},
  year = {2008},
  title = {Focus on Formative Feedback},
  journal = {Review of Educational Research},
  volume = {78},
  number = {1},
  pages = {153--189},
  doi = {10.3102/0034654307313795},
  url = {https://doi.org/10.3102/0034654307313795},
  citationcount = {4610},
  citationprovider = {OpenAlex},
  citationretrieved = {2026-10-23},
  citationurl = {https://api.openalex.org/works/https://doi.org/10.3102/0034654307313795?select=id,doi,title,cited_by_count,counts_by_year,authorships,primary_location,updated_date},
  journalimpactfactor = {8.6},
  jifyear = {2025},
  jifsource = {https://www.aera.net/Newsroom/AERA-Highlights-E-newsletter/AERA-Highlights-July-2026/AERA-Journals-Maintain-Strong-Standing-in-2026-Journal-Citation-Reports},
  jifverification = {Publisher or professional association},
  journalstanding = {Leading AERA review and research-synthesis journal in education.},
  authorinstitutions = {Florida State University, USA},
  affiliationsource = {https://journals.sagepub.com/doi/10.3102/0034654307313795 ; https://doi.org/10.3102/0034654307313795},
}

@article{Pearson2026,
  author  = {Pearson, Helen},
  title   = {How to stay smart in the age of {AI}: the science of critical thinking},
  journal = {Nature},
  year    = {2026},
  volume  = {657},
  number  = {8133},
  pages   = {873--875},
  doi     = {10.1038/d41586-026-02930-6},
  note    = {News Feature}
}

@article{kosmyna2025your,
  title={Your brain on ChatGPT: Accumulation of cognitive debt when using an AI assistant for essay writing task},
  author={Kosmyna, Nataliya and Hauptmann, Eugene and Yuan, Ye Tong and Situ, Jessica and Liao, Xian-Hao and Beresnitzky, Ashly Vivian and Braunstein, Iris and Maes, Pattie},
  journal={arXiv preprint arXiv:2506.08872},
  volume={4},
  pages={206},
  year={2025}
}

@article{ThoeniFryer2026,
  author  = {Thoeni, Andrew and Fryer, Luke K.},
  title   = {{AI} Chatbots in Higher Education: Comparing Expectations to Evidence},
  journal = {Computers in Human Behavior Reports},
  year    = {2026},
  volume  = {22},
  pages   = {101061},
  doi     = {10.1016/j.chbr.2026.101061},
  url     = {https://doi.org/10.1016/j.chbr.2026.101061}
}

@techreport{WangCasePrep2026,
  author      = {Wang, Dan J. and Jain, Neelam Modi and Burbano, Vanessa and Guzman, Jorge and Keum, Daniel and Kim, Soomi and Kogut, Bruce and Wright, Nataliya},
  title       = {How Assigned {AI} Use Before Class Shapes Active Student Engagement in Class},
  institution = {SSRN},
  type        = {Working Paper},
  number      = {7585318},
  year        = {2026},
  doi         = {10.2139/ssrn.7585318},
  url         = {https://papers.ssrn.com/sol3/papers.cfm?abstract_id=7585318},
  note        = {Version dated 7 October 2026; also available as arXiv:2610.10463}
}

@article{Burch2019Experiential,
  author  = {Burch, Gerald F. and Giambatista, Robert and Batchelor, John H. and Burch, Jana J. and Hoover, J. Duane and Heller, Nathan A.},
  title   = {A Meta-Analysis of the Relationship Between Experiential Learning and Learning Outcomes},
  journal = {Decision Sciences Journal of Innovative Education},
  year    = {2019},
  volume  = {17},
  number  = {3},
  pages   = {239--273},
  doi     = {10.1111/dsji.12188},
  url     = {https://doi.org/10.1111/dsji.12188}
}

@article{WiseJung2019Analytics,
  author  = {Wise, Alyssa Friend and Jung, Yeonji},
  title   = {Teaching with Analytics: Towards a Situated Model of Instructional Decision-Making},
  journal = {Journal of Learning Analytics},
  year    = {2019},
  volume  = {6},
  number  = {2},
  pages   = {53--69},
  doi     = {10.18608/jla.2019.62.4},
  url     = {https://doi.org/10.18608/jla.2019.62.4}
}

@techreport{PasqualeEtAl2025,
  author      = {Pasquale, Frank and Rand, David and Sengers, Phoebe and Smolka, Marcus and Stroock, Abraham and Thom-Levy, Julia},
  title       = {Subcommittee on Technological Change and Artificial Intelligence: Issue Summary and Core Questions},
  institution = {Cornell University, Committee on the Future of the American University},
  type        = {Committee report},
  year        = {2025},
  month       = nov,
  url         = {https://provost.cornell.edu/initiatives/future-of-the-american-university/publications-reports/technological-change-ai/},
  note        = {Accessed 6 November 2026}
}

@article{Kessler2018, title={Technology and the future of language teaching}, volume={51}, ISSN={1944-9720}, DOI={10.1111/flan.12318}, number={1}, journal={Foreign Language Annals}, publisher={Wiley}, author={Kessler, Greg}, year={2018}, month=Feb, pages={205--218} }

@article{MoonKessler2026, title={Exploring AI Integration in South Korean K-12 Education: Teachers’ Practices, Motivations, and Ethical Considerations}, volume={19}, ISSN={1916-4742}, DOI={10.5539/elt.v19n10p88}, number={10}, journal={English Language Teaching}, publisher={Canadian Center of Science and Education}, author={Moon, Hyunjoo and Kessler, Greg}, year={2026}, month=Sept, pages={88} }

@article{KepplerLiWu2022,
  author  = {Keppler, Samantha M. and Li, Jun and Wu, Di (Andrew)},
  title   = {Crowdfunding the Front Lines: An Empirical Study of Teacher-Driven School Improvement},
  journal = {Management Science},
  year    = {2022},
  volume  = {68},
  number  = {12},
  pages   = {8809--8828},
  doi     = {10.1287/mnsc.2022.4331},
  url     = {https://doi.org/10.1287/mnsc.2022.4331}
}

% =====================================================================
% METHODS
% =====================================================================

\clearpage
\section*{Methods}

\subsection*{Ethics}
We received ethical approval for all 19 participating institutions, either through a single institutional review board (IRB) reliance agreement, or through independent review by the institution's IRB or equivalent ethics body. The University of Michigan IRB served as the board of record for the single-IRB arrangement. Reliance agreements were established with Cornell University, Georgetown University, Indiana University, The Ohio State University, Purdue University, Southern Methodist University, Stevens Institute of Technology, The University of Texas at Dallas, the University of Virginia, and the University of Washington. The study was not federally funded, and use of a single IRB was not required. Independent approval was obtained from Carnegie Mellon University, The University of Hong Kong, the Indian Institute of Management Ahmedabad, Northwestern University, the University of Illinois Urbana-Champaign, the University of Pennsylvania, the University of South Carolina, and Yale University. 

\subsection*{Pre-registration}
The study was pre-registered on the Open Science Framework (osf.io) on 28 February 2025, and amended on 6 March 2025, 15 August 2025, and 24 August 2026.\footnote{The March 2025 amendment relaxed the eligibility criteria for historical comparison data. The August 2025 amendment detailed the General IV specification and allowed instructors to (i) enter additional rounds of randomization and (ii) introduce non-randomized AI innovations. The August 2026 amendment removed these two options, neither of which had been used by participating instructors, and introduced the Simple IV specification. Supplementary Materials S1 describes these amendments in more detail.}

\subsection*{RCT design and implementation}

\paragraph{Study population}
The study included applied STEM courses delivered in residential, hybrid, and online formats at participating institutions. Subject topics included statistics, probability, machine learning, optimization, simulation, data analytics and visualization, regression, forecasting, decision analysis, and modeling, with applications in economics, management, health care, supply chain and logistics, information technology, and finance. Overall, there were 30 undergraduate classes, and 25 graduate classes.

\paragraph{Recruitment and eligibility}
Recruitment proceeded in two stages. The research team first recruited site principal investigators (site PIs), who then recruited instructors teaching eligible courses at their institutions. Enrollment proceeded on a rolling basis, with courses grouped into cohorts by entry term. This structure allowed the study to account for term-specific factors, including changes in students' baseline AI usage. Eligible instructors were required to (i) have taught the course at least once before joining the experiment, (ii) have not yet systematically integrated AI into the curriculum, and (iii) have no planned major curricular changes during the experimental term other than the proposed AI innovations.\footnote{Some courses experienced unplanned curricular or instructor changes during the experiment and were subsequently excluded from the relevant analyses (see Analysis Sample).}

\paragraph{Participants}

The experiment involved 19 universities, 42 instructors, and 7,200 students. It comprised 55 classes and 67 experimental class-terms, where a \textit{class} is a course--instructor pair, and a \textit{class-term} is a course--instructor--term triple. The three entry cohorts, in Spring 2025, Fall 2025, and Spring 2026, comprised 15, 20, and 15 classes from newly randomized courses, respectively. Participation in previously assigned courses contributed an additional 3 class-terms in Fall 2025 and 14 in Spring 2026 (Fig.~\ref{fig-flow}b).

\paragraph{Randomization}

Before each semester, instructors submitted course titles, start and end dates, academic levels (undergraduate or graduate), modalities (residential, hybrid, or online), expected enrollments, and prior-year syllabi. We clustered courses by subject and academic level: most clusters comprised two or three classes, but some comprised only one, if the class is like no other. We block randomized assignment within cluster, ensuring that the fraction of each cluster assigned to the treatment group is as close as possible to one-half: e.g., all three-class clusters either have one or two treatment classes. Most professors contributed only one class to the experiment in a term, but some contributed two. When a professor contributed two classes in a term, we randomly assigned one to the treatment group, and the other to the control group. Further, if multiple instructors taught the same course at the same institution in the same term then we allocated their classes to the same experimental arm, with one ad hoc exception: for complicated reasons, two instructors taught a similar subject at the same institution but each to a different audience level (undergraduate vs. graduate) had to be in the same experimental cluster, and had to switch the treatment and control assignment in the second term they participated in the study. 

\paragraph{Equal access and student consent}

For ethical and instructional reasons, we randomized treatment at the course level, rather than the student level. We made sure all students enrolled in a course had equal access to any AI support or activity introduced in that course, regardless of whether they consented to sharing data for research use. Instructors de-identified the student-level data, dropping the learning and engagement scores for the students in the treatment class-terms who did not consent to sharing data. Across class terms, the consent rate had mean 79.4\%, and quartiles 60.0\%, 84.6\%, and 100\%.

\paragraph{Outcome metrics}
Before receiving their randomized assignment, instructors specified empirical measures for learning, engagement, and satisfaction. These measures had to be consistent across teaching terms, both before and during the experimental period. Learning measures were raw, uncurved assessment scores, such as points earned on quizzes or examinations (curved grades were ineligible because they obscure changes in class-level performance). Engagement measures included attendance, class participation, assignment completion, and logins and page views on the learning management system (LMS), such as Canvas or Blackboard. Satisfaction measures stemmed from the institution's anonymous course evaluations. Classes assigned to the same block randomization cluster were encouraged to choose similar measures, but consistency within a class over time took precedence over consistency across classes within a cluster.

\paragraph{AI innovation proposals}
Before randomization, instructors submitted a written AI innovation proposal for each participating course. Each proposal described (i) the planned use of AI and (ii) the rationale for that design, including how it may affect learning, engagement, and satisfaction. Proposals had to exceed students' ambient AI usage, incorporating the instructor's own design or customization: e.g., allowing ChatGPT on a homework assignment would not suffice, but creating a dedicated homework agent trained on the assignment and supporting course materials would.

\paragraph{Instructor support}
All instructors in both treatment and control arms received technical and AI-design support before and during the teaching term. Video conferences were held every two months throughout the academic year, covering model prompting, system instructions, principles of agentic AI, and the creation and deployment of agents on different platforms. A typical meeting also featured two to three instructors presenting and discussing their innovations. Two in-person consortium workshops, held in June 2025 and July 2026, provided more intensive training, demonstrations, and opportunities for instructors to exchange ideas.

Asynchronous resources included a teaching-with-AI tutorial\cite{Bray2025}, demonstration videos, and libraries of sample agent instructions. We also co-developed \textit{Faculty Twin} with Google's Rapid Innovation Team. This platform allowed instructors to configure and deploy AI agents through graphical user interfaces in a secure cloud environment compliant with the Family Educational Rights and Privacy Act (FERPA) and approved by IRB. 

\paragraph{Data collection}

Each instructor provided data from their classes after the start of the experiment, and from at least one term before the start of the experiment. As described in our pre-registered protocol, a pre-experimental benchmark was valid only if (i) it corresponded to the same course, instructor, and teaching modality, (ii) it was less than 10 years old, (iii) it was either from the most recent four years of the class or generated more than a year after the class's creation, (iv) it fell outside the institution's COVID-19 lockdown period, and (v) its data are sufficient to derive the measure being analyzed.

Instructors provided class-level summary statistics (mean, standard deviation, median, and quartiles) across all students, as well as de-identified individual scores for consenting or non-treatment students, per IRB requirement. All datasets were uploaded to a FERPA-compliant secure server managed by the University of Michigan's Information and Technology Services.

\paragraph{AI implementation}
Treatment classes were required to implement their proposed innovations in the assigned term and thereafter, and control classes were required to defer implementation for at least one term.\footnote{There was one exception. Two instructors who taught similar courses at one institution had to implement an AB/BA switch back design.} 
The next time a control instructor taught, they could (i) implement their AI proposal, (ii) not implement their AI proposal, or (iii) subject this decision to another round of randomization. In practice, every control instructor chose to implement their AI innovation in the following teaching term.

\paragraph{Attrition and exclusions}

Fig.~\ref{fig-flowchart} describes the filters we imposed upon our sample. We dropped two treatment class-terms from our ordinary least squares (OLS) regressions (but not the auxiliary instrumental-variables (IV) regressions presented in Extended Data Fig.~\ref{edfig-twelve}), because they did not meaningfully implement their proposals: one instructor could not implement the AI agents due to a lack of development time\footnote{Our ethics protocol requires instructors to test AI agents carefully before making them available to students. This instructor did not find responses from AI agents meet the desired accuracy level, but s/he lacked time to further refine the agents.}, and the other did not adequately inform students about the agents. And we excluded another class-term from the learning and engagement analyses because of a classification error: we didn't realize that there were two versions of the class (one for undergraduates and the other for graduates), so we failed to randomly assign the graduate version.

We excluded four additional class-terms following instructor changes and six more following major changes to the curriculum or assessment formats. We excluded eight additional class-terms from the satisfaction analysis, because their universities changed the course-evaluation survey questions.
And, finally, we excluded ten additional class-terms from the learning analysis, because their assessments were not AI-proof. 
Our definition of AI-proof assessments goes beyond prohibiting AI use through honor codes or written policies in the syllabus. We require verifiable procedures for ensuring that students cannot use AI when they take the assessments. Hence, all the AI-proof assessments in our analysis were proctored, in-person examinations, most of which were paper-based, and the rest strictly monitored, computer-based examinations that used surveillance software, such as LockDown Browser, or restricted-access computers at university-managed testing centers. We did not pre-register this last exclusion criterion, as we did not anticipate the problem of AI use on the assessments. But since real-time AI assistance can artificially enhance performance, we believe imposing this filter is necessary to properly capture true learning outcomes\cite{Deng2025,Weidlich2025}.

\paragraph{Analysis sample}
Restricting to experimental class-terms, our class-level sample comprises 44 class-terms for learning (4,768 enrolled students), 54 for engagement (5,637 students), and 47 for satisfaction  (5,020 students) (Fig.~\ref{fig-flowchart}), and our student-level sample comprises 44 class-terms for learning (4,768 students), 50 for engagement (5,225 students), and 47 for satisfaction  (5,020 students) (Fig.~\ref{fig-flowchart}).\footnote{
For four class-terms, engagement data were only available at the class level, but not at the student level.
} 
Including historical terms, our class-level sample comprises 101 class-terms for learning (10,471 students), 117 for engagement (11,394 students), and 102 for satisfaction (7,118 students), and our student-level sample comprises 101 class-terms for learning (9,202 students), 112 for engagement (9,658 students), and 102 for satisfaction (7,118 students).                 

\subsection*{Empirical specification}

\paragraph{Regression model}

Our regression equations have a difference-in-differences structure, comparing the student performance of the treatment group, relative to their historical average, with the student performance of the control group, relative to their historical average. A given equation characterizes one outcome (learning, engagement, or satisfaction) at one level of aggregation (student or class). For the student-level analyses, we use the following specification:
\[
z_{nti} = \alpha_n + \gamma_t + \beta_{0}\,\text{AI\_Innovation}_{nt} + \beta_{1}\,\text{Prior\_AI\_Innovation}_{nt} + \epsilon_{nti}.
\]

The dependent variable reports the \(z\)-score associated with student \(i\), teaching term \(t\), and class \(n\): \(z_{nti} = (x_{nti} - \mu_{n})/\sigma_{n}\), where \(x_{nti}\) denotes the student's scores winsorized at 3\% and 97\% levels, and \(\mu_{n}\) and \(\sigma_{n}\) denote the mean and standard deviation of these winsorized class-\(n\) scores, pooled across all of the class's students and teaching terms.\footnote{ 
	In the pre-registration, we committed to using \(z\)-scores in our analyses to account for heterogeneity across classes and outcome measures. We also committed to winsorizing the top and bottom 3\% of each empirical measure, by class, \emph{before} computing the moments \(\mu_{n}\) and \(\sigma_{n}\) that define these \(z\)-scores, to mitigate potential effects of extreme values.
} 
The regression explains these $z$ scores in terms of fixed effects $\alpha_n$ and $\gamma_{t}$, which account for class-level and term-level fluctuations, independent variables \(\text{AI\_Innovation}_{nt}\) and \( \text{Prior\_AI\_Innovation}_{nt} \), which account for the effects of AI, and statistical error $\epsilon_{nti}$, which accounts for the residual. Recall a class is defined as a course-instructor pair, so class fixed effects also capture teacher fixed effects. 

Both \(\text{AI\_Innovation}_{nt}\) and \( \text{Prior\_AI\_Innovation}_{nt} \) are binary variables: the former indicates that AI was introduced in period $t$, and the latter indicates that AI was introduced before period $t$. Specifically, \(\text{Prior\_AI\_Innovation}_{nt} = 1\) if and only if \(\text{AI\_Innovation}_{ns} = 1\) for some $s < t$, and \(\text{AI\_Innovation}_{nt} = 1\) if and only if the class was assigned to the treatment group in term $t$, or it was assigned to the control group in the last term prior to $t$ that the class was taught (recall that all control-group instructors implemented their AI innovations in the next teaching term).\footnote{
    There are two exceptions. First, two treatment instructors did not implement AI; for their classes, we set $\text{AI\_Innovation}_{nt} = 0$ for all $t$. Second, two instructors who taught similar courses at the same institution implemented an AB/BA switchback design; for their classes, we set \(\text{AI\_Innovation}_{nt}=1\) during their treatment terms and \(\text{Prior\_AI\_Innovation}_{nt}=0\) in every term.
} 
Note that the lagged effect, \(\beta_{1}\), could differ from the contemporaneous effect, \(\beta_{0}\), if there's a learning curve to AI adoption.

For the class-level analysis, we first derive class-level \(z\)-scores: \[
\begin{aligned}
\bar{z}_{nt} &= \frac{m_{nt} - \mu_{n}}{\sigma_{n}}, \\
\text{where} \quad \mu_{n} &= \frac{\sum_{t} c_{nt}\, m_{nt}}{\sum_{t} c_{nt}}, \\
\text{and} \quad \sigma_{n}^{2} &= \frac{\sum_{t} (c_{nt} - 1)\, s_{nt}^{2} + \sum_{t} c_{nt} \left( m_{nt} - \mu_{n} \right)^{2}}{\sum_{t} c_{nt} - 1}.
\end{aligned}
\] In the expressions above, \(m_{nt}\), \(s_{nt}\), and \(c_{nt}\)
correspond to the means, standard deviations, and student counts we
observe in our class-level data (i.e., aggregate statistics for all students enrolled in a class, regardless of consent for individual-level data).
Calculating the aggregate \(z\)-score in this manner makes the following
class-level regression equivalent to our student-level
regression (except for the 3\% winsorization and differing sample composition): \[
\sqrt{c_{nt}}\,\bar{z}_{nt} = \sqrt{c_{nt}}\,\alpha_n + \sqrt{c_{nt}}\,\gamma_t + \beta_{0}\sqrt{c_{nt}}\,\text{AI\_Innovation}_{nt} + \beta_{1}\sqrt{c_{nt}}\,\text{Prior\_AI\_Innovation}_{nt} + \nu_{nt}.
\] Note, the \(\sqrt{c_{nt}}\) scaling assigns equal weight to every
student, in alignment with our student-level analysis. 

\paragraph{Estimators}

For our primary analysis, reported in Fig.~\ref{fig-main}, we estimate AI-effect coefficients \(\beta_{1}\) and \(\beta_{0}\) with ordinary least squares (OLS), and compute confidence intervals with the CR2 bias-reduced cluster-robust variance estimator, clustering by class-term and using Student-$t$ critical values with Satterthwaite degrees of freedom\cite{BellMcCaffrey2002,ImbensKolesar2016}. For our auxiliary analyses, reported in Extended Data Fig.~\ref{edfig-twelve}, we consider two other estimators, and another set of confidence intervals. We construct the alternative confidence intervals by combining conventional standard errors, clustered by class-term, with critical values from a Student-$t$ distribution with $G-1$ degrees of freedom, where $G$ is the number of class-term clusters. And we derive the alternative estimates with instrumental variables (IV) regression.

The Simple IV specification instruments for
\(\text{AI\_Innovation}_{nt}\) with
\[
\text{Treatment}_{nt} + \text{Control}_{nt-1}.
\]
It instruments for
\(\text{Prior\_AI\_Innovation}_{nt}\) with
\[
\text{Prior\_Treatment}_{nt}
+ \text{Prior\_Control}_{nt-1}.
\]
Here \(\text{Treatment}_{nt}\), \(\text{Control}_{nt-1}\), \(\text{Prior\_Treatment}_{nt}\), and  \(\text{Prior\_Control}_{nt-1}\) are binary variables that respectively indicate a treatment assignment in the current teaching term, a control assignment in the previous teaching term, a treatment assignment prior to the current teaching term, and a control assignment prior to the previous teaching term (note that the \(t-1\) subscripts are a slight abuse of notation, because rather than the last term before \(t\), they refer to the last term before \(t\) that class \(n\) was taught). Note, since all control-group professors implemented their AI innovations in the subsequent teaching term, \(\text{Treatment}_{nt} + \text{Control}_{nt-1}\) equals \(\text{AI\_Innovation}_{nt}\) and \(\text{Prior\_AI\_Innovation}_{nt}\) equals \(\text{Prior\_Treatment}_{nt} + \text{Prior\_Control}_{nt-1}\), for all but the two non-compliant treatment-group class-terms. Indeed, this IV regression would be equivalent to our OLS regression if it weren't for the two treated instructors who did not implement their AI proposals (we exclude their classes for the OLS regression, which cannot accommodate non-compliance, but not for the IV regressions, which can accommodate non-compliance).

The General IV specification instruments for
\(\text{AI\_Innovation}_{nt}\) and
\(\text{Prior\_AI\_Innovation}_{nt}\)
with the following four instruments:
\[
\begin{aligned}
\frac{\text{Treatment}_{nt}}
     {\text{Total\_Assignments}_{n}},
&\qquad
\frac{\text{Control}_{nt}}
     {\text{Total\_Assignments}_{n}}, \\[6pt]
\frac{\text{Prior\_Treatment}_{nt}}
     {\text{Total\_Assignments}_{n}},
&\qquad
\frac{\text{Prior\_Control}_{nt}}
     {\text{Total\_Assignments}_{n}}.
\end{aligned}
\]

The numerator variables are defined analogously to those in the Simple IV specification, but with \(\text{Control}_{nt}\) and \(\text{Prior\_Control}_{nt}\) subscripted with $t$, rather than $t-1$. The denominator, \(\text{Total\_Assignments}_{n} \equiv \sum_s (\text{Treatment}_{ns} + \text{Control}_{ns})\), is the total number of experimental batches that class \(n\) enrolled in.\footnote{
    For the two instructors implementing the AB/BA switchback design, we set the Simple IV instruments to \((\text{Treatment}_{nt},0)\), and the General IV instruments to \((\text{Treatment}_{nt},\text{Control}_{nt},0,0)\), in the order given above.
}
We discuss these IV specifications in more depth in Supplementary Materials S1.

Extended Data Fig.~\ref{edfig-twelve} reports all 12 combinations of the three estimators, two aggregation levels, and two inference methods. All tests were two-sided, with 90\% and 95\% confidence intervals.

\subsection*{Student survey textual analysis}

We classified the responses to ``What did you like about the AI learning aids used in this class?'' with the following procedure: First, we exclude responses that are empty, non-informative (e.g., ``Good.'' without further elaboration), or strictly concerning software functionality (e.g., ``the interface is simple''). This filter limited the sample to 1,414 responses. Second, a researcher identified six prominent themes after reading a random sample of 200 responses. Independently, a large language model (\textit{gpt-6-astra}) identified six themes after reading all responses. The human and AI themes overlapped significantly: combining the lists yielded seven unique themes. We then instructed the LLM to label all responses according to these seven themes. A given response could receive multiple labels. Last, we confirmed the accuracy of the AI's theme designations by comparing them to the themes specified by a human researcher for a random 200 responses: the human and AI agreed for 95.9\% of labeling decisions. 

The responses to the other two open-ended questions ``What did you dislike about the AI learning aids used in this class?'' and ``How can we improve the AI learning aids used in this class?'' overlapped significantly.\footnote{
    A student would usually either leave their response to the second question blank, or they would repeat what they responded to the first question.
} 
Thus, we combined the responses to these questions, and analyzed the pooled collection with the procedure described above. Again, the human labeler and the LLM identified overlapping themes (ten in total), and agreed for 95.8\% of labeling decisions. 
 
\subsection*{Instructor interview}

Following the study, we performed a semi-structured interview with each of the 25 instructors who designed AI innovations for their classes and whose data were included in the analysis sample. Each interview lasted about 25 minutes. 24 instructors were interviewed over zoom online meetings and one interviewed via email due to scheduling conflict.  
Our semi-structured interviews explored the following questions: 

\textit{Part 1 - What were you trying to do with AI?}
Sample prompt: Could you briefly describe your course and the AI intervention(s) you implemented? What teaching problem, pain point, or opportunity were you trying to address, and why did you choose this particular use of AI?

\textit{Part 2 - Why did you think AI would help with that problem?}
Sample prompt: What did AI make possible or easier in this setting? Could you have addressed the same problem without AI, or was there something about AI that made it particularly well suited to what you were trying to do?

\textit{Part 3 - What actually happened when you implemented it?}
Sample prompt: How did students actually use or interact with the AI? Did they use it in the way you expected? Was there anything that worked differently than expected, surprised you, or caused you to change the intervention during the semester?

\paragraph{Interview analysis}

We conducted a thematic analysis of interview notes employing an inductive approach to identify patterns (themes) within the data. Because instructors' objectives and how AI benefited were closely connected, we analyzed responses to Parts 1 and 2 jointly. Responses to Part 3 were analyzed separately for implementation experiences, including actual student use, comparisons with expectations, surprises and subsequent adjustments. 
The research team first manually reviewed the interview notes line by line to identify recurring low-level codes, which were then grouped into higher-level themes based on conceptual similarity, and were revised iteratively through peer debriefing within the team of five primary researchers until they reached consensus. The resulting framework comprised seven themes for objectives and how AI benefited, and five for implementation experiences. Each interview was counted once within each theme. An interview could contribute to multiple themes.

\subsection*{Reporting summary}
Further information is available in the Nature Portfolio Reporting Summary linked to this article. 

\subsection*{Data availability}
All data used in this research are stored on a secure server managed by the University of Michigan Information and Technology Services (ITS), in accordance with Institutional Review Board requirements. Restricted-use research data will be made available to qualified researchers with appropriate IRB approval and upon execution of a data use or memorandum of understanding (MOU) agreement, as applicable.

\section*{Code availability}
The quantitative analyses were conducted in R. The analysis code is stored on the secure server managed by the University of Michigan Information and Technology Services (ITS) and is available to editors and referees upon request.

% =====================================================================
% METHODS REFERENCES: at submission, the Methods-only references
% above) are listed here, continuing the main numbering.
% =====================================================================

% =====================================================================
% END MATTER
% =====================================================================
\section*{Funding}
Research reported in this manuscript was supported by discretionary research funds from Northwestern University and the University of Michigan. 

\section*{Acknowledgments}
The authors acknowledge the University of Michigan Health Sciences and Behavioral Sciences Institutional Review Board (IRB-HSBS) for serving as the Board-of-Record for the relying institutions in this multisite study. We particularly thank Corey Zolondek and the IRB-HSBS staff for their leadership and administrative coordination of the single-IRB review, oversight, and reliance arrangements.

The authors thank Gina Fratarcangeli and Google's Rapid Innovation Team for its collaboration in co-developing the Faculty Twin AI platform to support the research. The authors also thank the University of Michigan Information and Technology Services (ITS) for providing staff and computing support for the deployment and management of the Faculty Twin platform in a secure, FERPA-compliant environment, and Kenneth Moore in particular for his technical assistance.

\section*{Author contributions}
J.~Li, R.~Bray, S.~Martin, N.~Van~den~Steen and A.~Wu are the primary research team and share first authorship.

J.~Li oversaw the study: she led instructor recruitment, led ethics protocol development, and contributed to study design, data collection, statistical analyses, Faculty Twin platform development and manuscript preparation. R.~Bray conceived the study and led the study design and statistical analyses and contributed to data collection and manuscript preparation. S.~Martin led AI innovation, AI training, and in-person workshops and contributed to study design, instructor recruitment and manuscript preparation. N.~Van~den~Steen led RCT protocol implementation, data collection, instructor interviews and qualitative data analysis, and contributed to study design and manuscript preparation. A.~Wu led the development of the Faculty Twin platform, its deployment in a FERPA-compliant environment, subsequent upgrades and instructor training, and contributed to study design, instructor recruitment and manuscript preparation.  X.~Ni led data audits and sample creation and contributed to data collection and statistical analyses. 

The site principal investigators were J.~Li, V.~O.~Babich, H.~Bastani, C.~Blanco, S.~Chen, Y.~Cui, P.~Feldman, D.~Hurtado-Lange, Z.~Jiang, Z.~Lian, M.~Lu, M.~Park, A.~Qi, D.~Roy, T.~Tan, N.~Tereyagoglu, Y.~Wang, O.~Wu and Z.~Yang. They recruited instructors, oversaw ethical and research protocols at corresponding sites. 

N.~Van~den~Steen, A.~Wu, A.~Anderer, E.~Astashkina, C.~Blanco, C.~Chen, S.~Chen, B.~Collier, Y.~Cui, P.~Feldman, B.~Flicker, Y.~Hu, D.~Hurtado-Lange, J.~Jiang, Z.~Jiang, M.~K{\"o}ksalan, Z.~Lian, M.~Lu, L.~Meng, K.~Moon, M.~K.~Newsom, M.~Park, C.~Parker, A.~Qi, M.~Qi, D.~Roy, N.~Sharma, A.~Sungu, T.~Tan, N.~Tereyagoglu, S.~Venkataraman, G.~Wang, Y.~Wang, Z.~Yang, Q.~Yu, J.~Zhang, W.~Zhang and R.~Zhu recruited participants, designed AI interventions and de-identified student level data. J.-P.~Baldwin, O.~Celebi, B.~Han, I.~Kirci, H.~Liang and H.~Tyagi recruited participants. S.~Yu designed AI interventions.

H. Bastani, C. Blanco, P. Feldman, B. Flicker, K. Moon, C. Parker, D. Roy, A. Sungu, T. Tan, N. Tereyagoglu, S. Venkataranman, O. Wu, and W. Zhang provided feedback and editorial input on the manuscript.

\section*{Competing interests}
The authors declare no competing interests.

\clearpage
\section*{Figures and tables}

%%%%%%%%% Figure showing geographical distribution 

\clearpage
\begin{figure}[p]
\centering
\includegraphics[width=\linewidth]{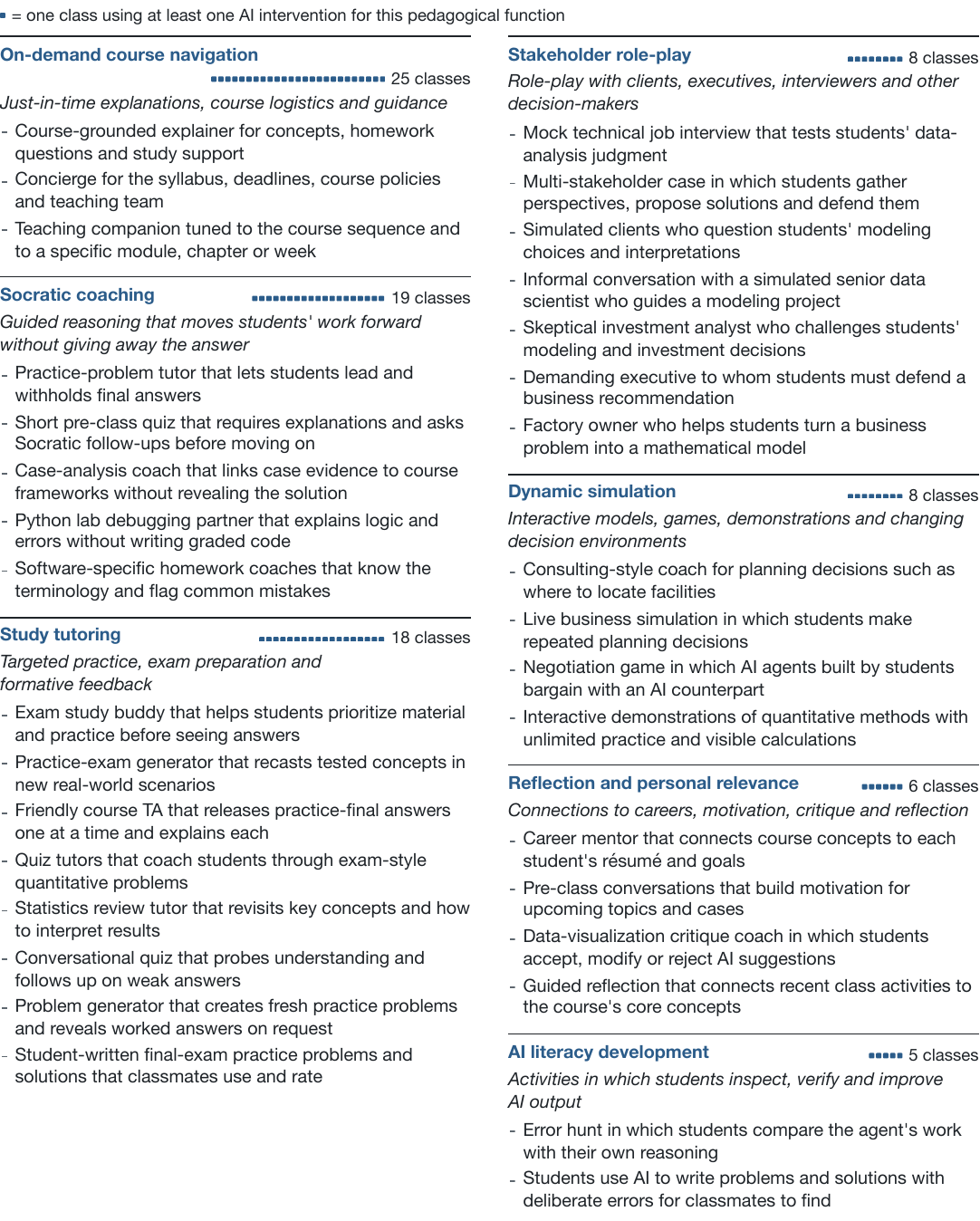}
\caption{\textbf{Instructor-designed AI innovations by pedagogical function.} Legend on the next page.}
\label{fig-interventions}
\end{figure}
\begin{figure}[p]
\ContinuedFloat
\caption*{\textbf{Figure~\ref{fig-interventions} (continued).} We characterized AI interventions using the agent prompts used by the instructors, post-implementation instructor surveys, and pre-registered innovation proposals. We restricted this characterization to classes that received treatment and were included in at least one class-level regression analysis, retaining only interventions involving direct student interaction. These restrictions yielded a sample of 30 classes. We combined interventions that created substantively the same student experience, yielding 33 types of distinct student-facing GenAI interventions. Subsequently, we identified seven recurring pedagogical functions and manually assigned each intervention one primary and, where clearly supported, one secondary function. Interventions are listed under their primary pedagogical function. Squares count classes with at least one intervention serving the corresponding function, whether primary or secondary, including interventions listed under another primary function. Each class contributes at most once to a given function's count, even when several interventions in that class serve that function. This yielded 89 unique class–function pairs across 30 classes, averaging 3.0 distinct pedagogical functions per class.}
\end{figure}

\clearpage

\clearpage

\clearpage
\clearpage
\begin{figure}[p]
\centering
\includegraphics[
  width=\linewidth,
  height=\textheight,
  keepaspectratio
]{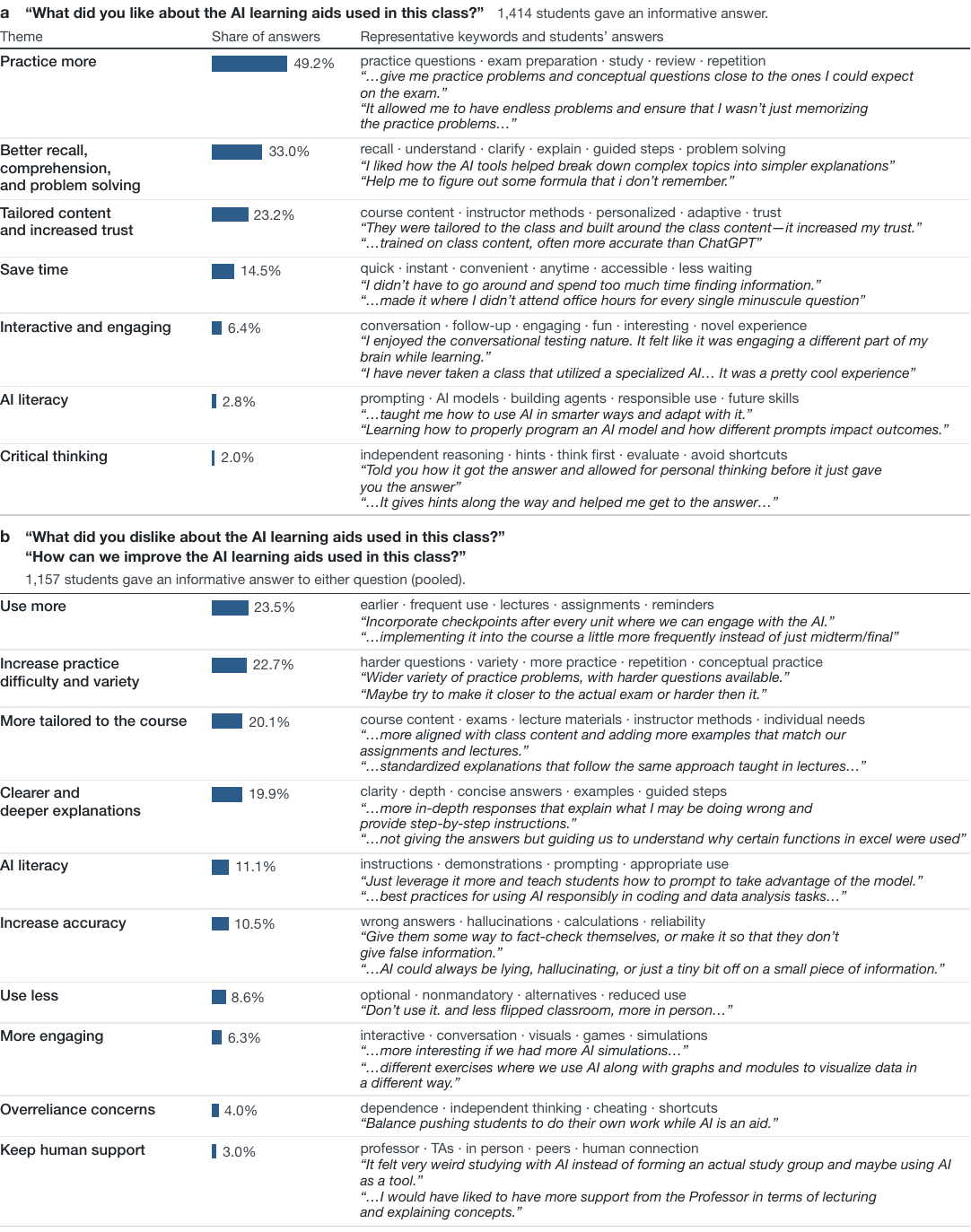}
\caption{\textbf{Student Survey Responses - Open-Ended Questions.}
Legend on the next page.}
\label{fig-survey}
\end{figure}

\clearpage
\begin{figure}[p]
\ContinuedFloat
\caption*{\textbf{Figure~\ref{fig-survey} (continued).}
Panel \textbf{a} summarizes themes in students' responses to the open-ended question ``What did you like about the AI learning aids used in this class?''
Panel \textbf{b} summarizes themes in responses to ``What did you dislike about the AI learning aids used in this class?'' and ``How can we improve the AI learning aids?''
Responses to these two questions overlapped substantially and were therefore analyzed together.
Themes were identified independently by a human researcher and a large language model (LLM), then manually reviewed and refined.
See Methods and Extended Data Tables~\ref{tab:survey-v3-1}
and~\ref{tab:survey-v3-2} for details.}
\end{figure}

\clearpage

% =====================================================================
% EXTENDED DATA
% =====================================================================
% =====================================================================
% SUPPLEMENTARY INFORMATION 
% =====================================================================
\clearpage
\section*{Supplementary materials}

\subsection*{S1. Pre-registration and Estimators}

We pre-registered our study on February 28, 2025, and amended it on March 6, 2025, August 15, 2025, and August 24, 2026. The March 2025 amendment relaxed an exclusion restriction. We initially planned on dropping data from the first two years a course was taught, to avoid classes' initial growing periods with potentially volatile student enrollments. We decided to be less restrictive, after learning it is practically difficult to enroll classes in the study under this requirement. 

The August 2025 amendment provided the first detailed characterization of our estimator. The original version provided a high-level description of the regression model and the difference-in-differences specifications, but it did not explain how we would estimate its coefficients. The August 2025 amendment rectified this omission, describing the General IV specification in detail. In conjunction, we also made a few minor adjustments to the statistical specification, including making the winsorization thresholds symmetric (from 3\% and 99\% to 3\% and 97\%), and replacing academic-year fixed effects with teaching-term fixed effects. 

The August 2025 amendment also made one modification to our experimental design, in response to our in person workshop in June 2025, where the consensus was that, due to the anticipated proliferation of AI usage across universities, it would soon be difficult to enroll new classes that had not yet incorporated any AI. In response, we updated the protocol to provide instructors options to (i) participate in in multiple experimental batches by subjecting follow-up AI proposals to randomization, and (ii) incorporate other non-randomized AI innovations (i.e., new AI features added outside of the experimental context). All modifications were made before analyzing the data.

The General IV specification instruments for $\text{AI\_Innovation}_{nt}$ and $\text{Prior\_AI\_Innovation}_{nt}$ using the following four instruments: $\frac{\text{Treatment}_{nt}}{\text{Total\_Assignments}_{n}}$, $\frac{\text{Control}_{nt}}{\text{Total\_Assignments}_{n}}$, $\frac{\text{Prior\_Treatment}_{nt}}{\text{Total\_Assignments}_{n}}$, and $\frac{\text{Prior\_Control}_{nt}}{\text{Total\_Assignments}_{n}}$. 
This specification divides the independent variables by \(\text{Total\_Assignments}_{n}\) to correct for instructors self-selecting into multiple rounds of experimentation. 

The August 2026 amendment reverted the August 2025 design modification. No instructor exercised either option, thereby making the modification moot. Correspondingly, the August 2026 amendment introduced the Simple IV specification, which is essentially the General IV specification without the August 2025 modification. Specifically, switching to Simple IV drops the \(\text{Total\_Assignments}_{n}\) denominators; because a class participates in only one round of experimentation, this variable is now mechanically one. The Simple IV specification also replaces the four instruments \(\text{Treatment}_{nt}\), \(\text{Control}_{nt}\), \(\text{Prior\_Treatment}_{nt}\), and \(\text{Prior\_Control}_{nt}\) with the two instruments \(\text{Treatment}_{nt} + \text{Control}_{nt-1}\) and \(\text{Prior\_Treatment}_{nt} + \text{Prior\_Control}_{nt-1}\). The rationale for combining \(\text{Treatment}_{nt}\) and \(\text{Control}_{nt-1}\) in the Simple IV is that these variables have essentially the same effect without the two modifications, namely making the course implement AI for the first time. 
Likewise for combining \(\text{Prior\_Treatment}_{nt}\) and \(\text{Prior\_Control}_{nt-1}\): these variables are almost perfectly co-linear with \(\text{Prior\_AI\_Innovation}_{nt}\). 

Further, we did not pre-register our OLS regression, but that specification is almost equivalent to the Simple IV specification, differing only in its treatment of the two non-compliant treatment-group class-terms, which, in turn, is similar to the General IV specification, differing only in its construction of the instruments. All three specification yield similar results. We did not pre-register our class-level regressions. Our pre-registration specifies that we use individual-level data, but the release of these student records are sometimes affected by low consent rates. Importantly, we can conduct almost the exact analysis we had pre-registered with our class-level data, which includes all students, consenting and not. Indeed, the only step in our pre-registered analysis that we cannot perform with our class-level aggregates is the 3\% winsorization of the distribution's tails. Other than that, our class-level regression is mechanically equivalent to the student-level regression that we had pre-registered.

\clearpage
\section*{Supplementary figures and tables}
\renewcommand{\thefigure}{ED\arabic{figure}}
\renewcommand{\thetable}{ED\arabic{table}}
\renewcommand{\theHfigure}{ED\arabic{figure}}
\renewcommand{\theHtable}{ED\arabic{table}}
\setcounter{figure}{0}
\setcounter{table}{0}

\begin{figure*}[!hp]
  \centering
  \includegraphics[width=\textwidth]{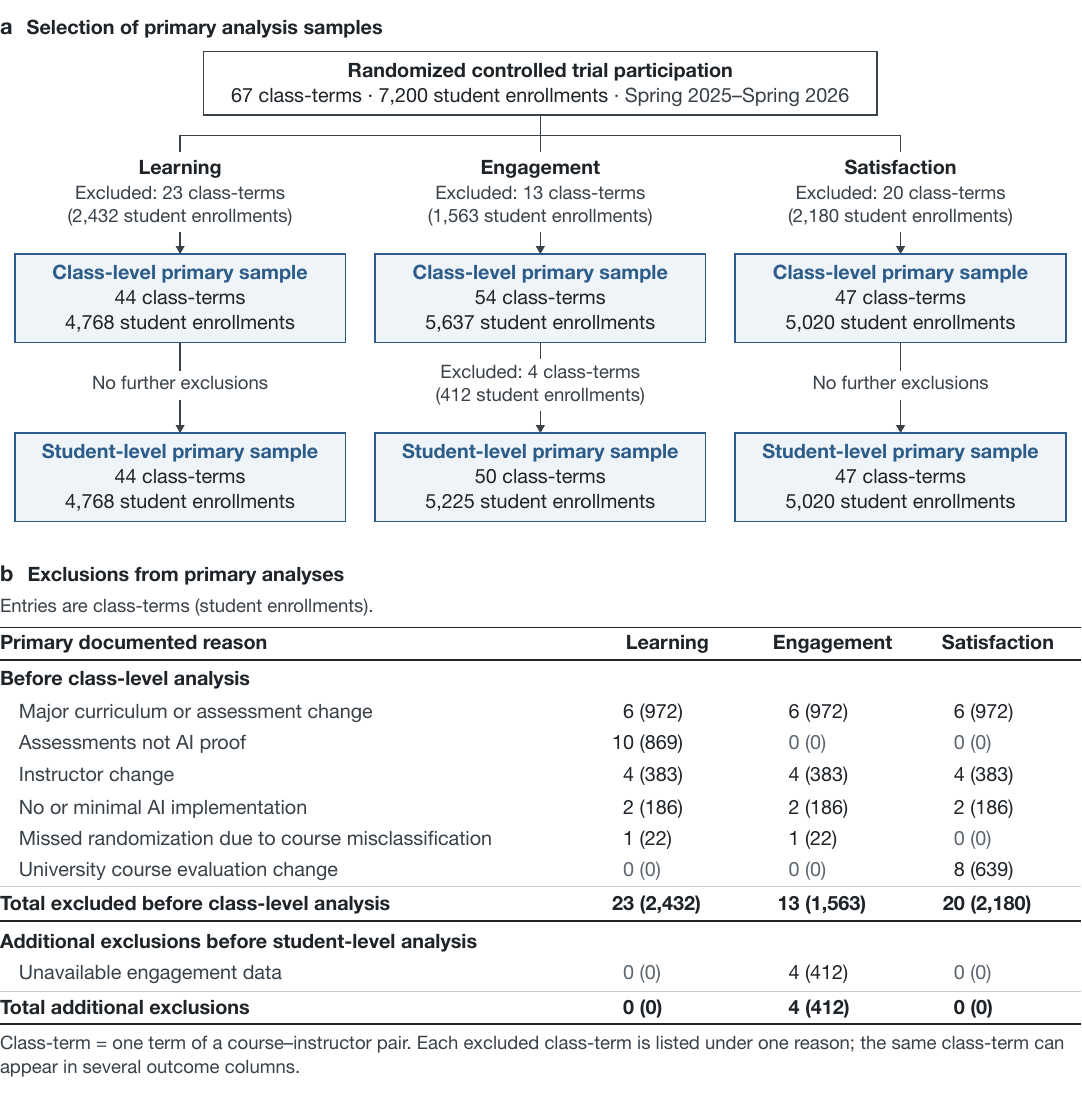}
  \caption{Inclusion and exclusion criteria for the analysis sample}
  \label{fig-flowchart}
\end{figure*}

\begin{figure}[!htbp]
\centering
\includegraphics[width=\linewidth]{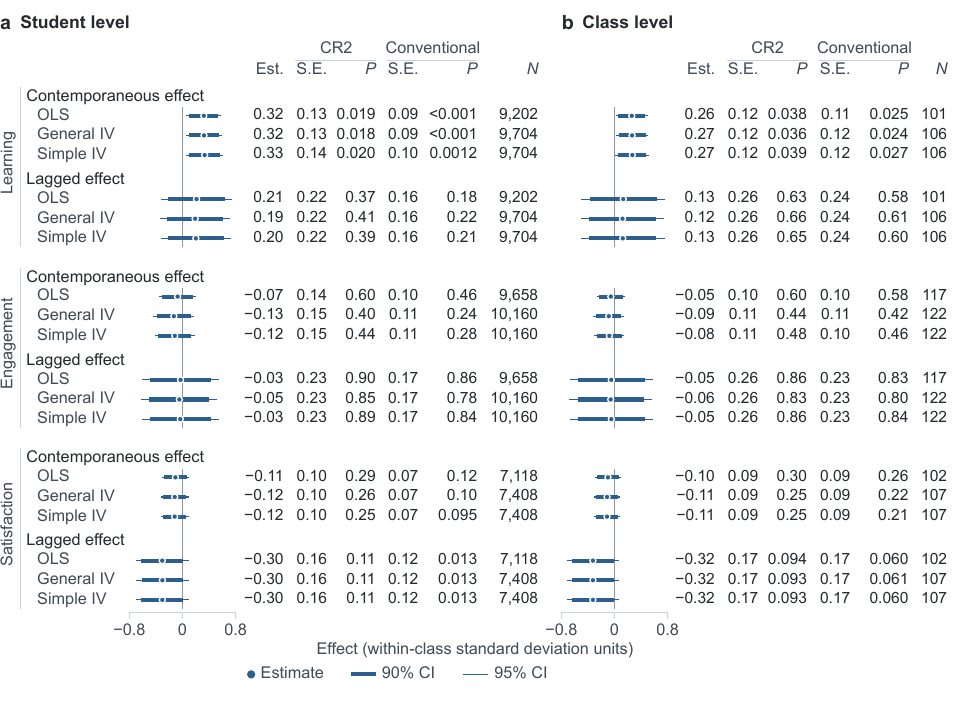}
\caption{\textbf{Estimates across alternative empirical specifications.} Points show effect estimates for student learning, engagement, and satisfaction, expressed in within-class standard deviation units; thick bars show 90\% and thin lines 95\% confidence intervals. Panel \textbf{a} reports student-level estimates; panel \textbf{b} reports class-level estimates. Both panels report contemporaneous effects in the first term of AI implementation and lagged effects associated with having implemented AI previously. Only three class-terms in the eligible analysis sample contributed observations from second or third AI implementations, limiting the precision of lagged estimates. For each outcome, twelve specifications combine three estimators (ordinary least squares [OLS], the general instrumental-variable [IV] estimator, and the simple IV estimator), two aggregation levels (student and class), and two confidence-interval methods (conventional and CR2). OLS analyses exclude the two treatment-assigned class-terms that did not implement their innovations; both IV analyses include them. The IV specifications are described in Methods, Empirical Specifications. Student-level aggregation was pre-registered, but restricting learning and engagement data to consenting students may introduce consent bias (average class-level consent rate 79.4\%, with a median of 84.6\%). Class-level learning and engagement summaries include all enrolled students, including those who did not consent to individual data use; satisfaction summaries reflect anonymous course-evaluation respondents that do not require consent. Both interval methods use Student-$t$ distributions and standard errors clustered by class-term. Conventional intervals use $G-1$ degrees of freedom, where $G$ is the number of class-term clusters; CR2 intervals use bias-reduced cluster-robust variance estimates and Satterthwaite degrees of freedom.}
\label{edfig-twelve}
\end{figure}

\clearpage
\begingroup
\ifdefined\nolinenumbers\nolinenumbers\fi
\setstretch{1}
\fontsize{12}{13}\selectfont
\setlength{\parskip}{0pt}
\setlength{\parindent}{0pt}
\setlength{\tabcolsep}{2pt}
\setlength{\LTleft}{0pt plus 1fill}
\setlength{\LTright}{0pt plus 1fill}
\setlength{\LTpre}{0pt}
\setlength{\LTpost}{5pt}
\setlength{\LTcapwidth}{\linewidth}
\setlength{\surveyUsableWidth}{\dimexpr\linewidth-6\tabcolsep\relax}
\setlength{\dashlinedash}{3pt}
\setlength{\dashlinegap}{2pt}
\setlength{\arrayrulewidth}{0.35pt}
\renewcommand{\arraystretch}{1}
\captionsetup[longtable]{font=normalsize,labelfont=bf,labelsep=period,
  justification=raggedright,singlelinecheck=false,skip=5pt}

\begin{longtable}{@{}
  >{\RaggedRight\arraybackslash}p{\surveyThemeShare\surveyUsableWidth}
  >{\Centering\arraybackslash}p{\surveyCountShare\surveyUsableWidth}
  >{\RaggedRight\arraybackslash}p{\surveyKeywordShare\surveyUsableWidth}
  >{\RaggedRight\arraybackslash}p{\surveyQuoteShare\surveyUsableWidth}@{}}
\caption{\textbf{Students' positive perceptions of AI learning components}}\label{tab:survey-v3-1}\\
\toprule
\textbf{Theme}&\textbf{N (\%)}&\textbf{Representative keywords}&\textbf{Sample quotes}\\
\midrule\endfirsthead
\multicolumn{4}{@{}l}{\textbf{Table \thetable{} (continued)}}\\
\toprule
\textbf{Theme}&\textbf{N (\%)}&\textbf{Representative keywords}&\textbf{Sample quotes}\\
\midrule\endhead
\midrule\multicolumn{4}{r@{}}{\textit{Continued on next page}}\\\endfoot
\bottomrule\endlastfoot
Practice more & 696\newline (49.2\%) & Practice questions; exam preparation; study; review; repetition & [1] ...give me \textbf{practice problems} and conceptual questions \textbf{close to} the ones I could expect on the exam.\newline [2] It allowed me to have \textbf{endless problems }and ensure that I wasn't just memorizing the practice problems \textbf{but }\textbf{actually knew}\textbf{ the steps} in a new problem \\[3pt]
\hdashline\noalign{\vskip 3pt}
Better recall, comprehension, and problem solving & 466\newline (33.0\%) & Recall; understand; clarify; explain; guided steps; problem solving & [3] Help me to figure out some formula that i don't \textbf{remember}.\newline [4] I liked how the AI tools helped \textbf{break down complex }topics into simpler explanations\newline [5] I liked that the AI tools gave \textbf{clear, step-by-step explanations} for Excel functions and quantitative concepts, which made the material much \textbf{easier to understand}. They also helped me check my work quickly and \textbf{reinforced what we learned} in class \ldots{}\newline [6] I liked that the AI learning aids made it \textbf{easier to understand complex} coding and modeling concepts, especially when working with large datasets and machine learning models. \\[3pt]
\hdashline\noalign{\vskip 3pt}
Tailored content and increased trust & 328\newline (23.2\%) & Course content; instructor methods; personalized; adaptive; trust & [7] They were \textbf{tailored to} the class and \textbf{built }\textbf{around} the class content---it increased my \textbf{trust}.\newline [8] .. referred to the \textbf{correct vocabulary }that was in the course...\newline [9] It helped me understand in \textbf{the same learning approach }that the professor used. \newline [10]\ldots{} \textbf{trained on class content}, often \textbf{more accurate }than ChatGPT \\[3pt]
\hdashline\noalign{\vskip 3pt}
Save time & 205\newline (14.5\%) & Quick; instant; convenient; anytime; accessible; less waiting & [11]... the AI helped me understand the conceptual things well and made it where I \textbf{didn't attend office hours }for every \textbf{single minuscule question}\newline [12] I didn't have to \textbf{go around and spend too much time }finding information.\newline [13] I also didn't have to \textbf{wait a long time }like for the professor to respond to an \textbf{email }to get a basic explanation.  \\[3pt]
\noalign{\vskip 3pt}
Interactive and engaging & 90\newline (6.4\%) & Conversation; follow-up; engaging; fun; interesting; novel experience & [14] I enjoyed the \textbf{conversational} testing nature. It felt like it was \textbf{engaging a different part of my brain }while learning.\newline [15] I have \textbf{never taken }a class that utilized a specialized AI in the way Professor Nils did. It was a \textbf{pretty cool}\textbf{ experience} \\[3pt]
\hdashline\noalign{\vskip 3pt}
AI literacy & 40\newline (2.8\%) & Prompting; AI models; building agents; responsible use; future skills & [16] Learning how to \textbf{properly program an AI model} and how different \textbf{prompts} impact outcomes.\newline [17] ... taught me how to \textbf{use AI in smarter ways} and adapt with it.\newline [18] It made it relatable to what we are likely to see in the future especially with \textbf{programming our own agent} \\[3pt]
\hdashline\noalign{\vskip 3pt}
Critical thinking & 28\newline (2.0\%) & Independent reasoning; hints; think first; evaluate; avoid shortcuts & [19] Gave me a chance to \textbf{think about what facts} I needed to apply, as opposed to being given all the facts...\newline [20] Told you how it got the answer and allowed for \textbf{personal thinking before} it just gave you the answer\newline [21] I like how this AI learning tool \textbf{doesn't give answers }to my hw questions right away. It gives hints along the way and helped me \textbf{get to the answer }-- why I think is a much more \textbf{effective learning process} than asking Chatgpt to give the final answer. \\[3pt]
\end{longtable}
\noindent\textit{Note.} In total, 2,019 out of 4,898 students enrolled in treatment classes responded to the survey and 1,689 students responded to the free-text question ``What did you like about the AI learning aids used in this class?'' Among them, 1,414 were informative responses. Non-informative responses included non-use, no liked aspect, or no substantive information (such as ``I liked the AI aid''). Themes were identified through human and LLM reading of the responses with the following process. First, human reader selected a random sample of 200 responses and identified the prominent themes. LLM independently read all responses and also identified prominent themes. Human and LLM both initially identified the first 5 of the 7 themes listed above. The human reader independently identified critical thinking, whereas LLM independently identified AI literacy. We thus arrived at 7 themes in total. LLM then labeled all responses according to the 7 themes. To verify accuracy, the human reader labeled the 200 randomly selected responses and calculated the human-LLM classification agreement rate of 95.9\%. Theme labels were not mutually exclusive; a response could contain more than one theme. Therefore, frequencies and percentages do not sum to the sample total or 100\%. Percentages in the table are calculated as percentages of the 1,414 informative responses.\par

\clearpage

\begin{longtable}{@{}>{\RaggedRight\arraybackslash}p{\surveyThemeShare\surveyUsableWidth}>{\Centering\arraybackslash}p{\surveyCountShare\surveyUsableWidth}>{\RaggedRight\arraybackslash}p{\surveyKeywordShare\surveyUsableWidth}>{\RaggedRight\arraybackslash}p{\surveyQuoteShare\surveyUsableWidth}@{}}
	\caption{\textbf{Students' criticisms and suggestions for improving AI learning aids}}\label{tab:survey-v3-2}\\
	\toprule
	\textbf{Theme} & \textbf{N (\%)} & \textbf{Representative keywords} & \textbf{Sample quotes} \\
	\midrule
	\endfirsthead
	\multicolumn{4}{@{}l}{\textbf{Table \thetable{} (continued)}}\\[3pt]
	\toprule
	\textbf{Theme} & \textbf{N (\%)} & \textbf{Representative keywords} & \textbf{Sample quotes} \\
	\midrule
	\endhead
	\midrule
	\multicolumn{4}{r@{}}{\textit{Continued on next page}}\\
	\endfoot
	\bottomrule
	\endlastfoot
	Use more
	& 272\newline (23.5\%)
	& Earlier; frequent use; lectures; assignments; reminders
	& [1] ... implementing it into the course a little \textbf{more frequently} instead of just midterm/final [2] Incorporate checkpoints \textbf{after every unit} where we can engage with the AI. \\[3pt]
	\hdashline
	\noalign{\vskip 3pt}
	Increase practice difficulty and variety
	& 263\newline (22.7\%)
	& Harder questions; variety; more practice; repetition; conceptual practice
	& [3] \textbf{Wider variety} of practice problems, with \textbf{harder questions} available. [4] Maybe try to make it closer to the actual exam or \textbf{harder then it}. [5] Help AI create \textbf{more challenging ideas}. \\[3pt]
	\hdashline
	\noalign{\vskip 3pt}
	More tailored to the course
	& 232\newline (20.1\%)
	& Course content; exams; lecture materials; instructor methods; individual needs
	& [6] ...more \textbf{aligned} with class content and adding more examples that \textbf{match} our assignments and lectures. [7] ...standardized explanations that follow the \textbf{same approach taught} in lectures... \textbf{tailored to} our specific assignments. \\[3pt]
	\hdashline
	\noalign{\vskip 3pt}
	Clearer and deeper explanations
	& 230\newline (19.9\%)
	& Clarity; depth; concise answers; examples; guided steps
	& [8] ...\textbf{more in-depth} responses that explain what I may be \textbf{doing wrong} and provide \textbf{step-by-step} instructions. [9] Using it to aid in the excel assignments, \textbf{not giving the answers} but guiding us to \textbf{understand why} certain functions in excel were used \\[3pt]
	\hdashline
	\noalign{\vskip 3pt}
	AI literacy
	& 129\newline (11.1\%)
	& Instructions; demonstrations; prompting; appropriate use
	& [10] One possible improvement would be to provide more \textbf{structured guidance or recommended use cases} for AI tools within the course, such as examples of \textbf{effective prompts} or \textbf{best practices} for using AI responsibly in coding and data analysis tasks. [11] Just leverage it more and teach students \textbf{how to prompt} to take advantage of the model. \\[3pt]
	\hdashline
	\noalign{\vskip 3pt}
	Increase accuracy
	& 122\newline (10.5\%)
	& Wrong answers; hallucinations; calculations; reliability
	& [12] Give them some way to \textbf{fact-check themselves}, or make it so that they don't give \textbf{false information}. [13] ...AI could always be lying, \textbf{hallucinating,} or just a tiny bit off on a small piece of information. \\[3pt]
	\hdashline
	\noalign{\vskip 3pt}
	Use less
	& 99\newline (8.6\%)
	& Optional; nonmandatory; alternatives; reduced use
	& [14] \textbf{Don't use it}. and less flipped classroom, more \textbf{in person} \ldots{}Better learning experience when following along in person instead of hearing video \\[3pt]
	\noalign{\vskip 3pt}
	More engaging
	& 73\newline (6.3\%)
	& Interactive; conversation; visuals; games; simulations
	& [15] ...more interesting if we had more AI \textbf{simulations} ... [16] ...different exercises where we use AI along with graphs and modules to \textbf{visualize data} in a \textbf{different way}. \\[3pt]
	\hdashline
	\noalign{\vskip 3pt}
	Overreliance concerns
	& 46\newline (4.0\%)
	& Dependence; independent thinking; cheating; shortcuts
	& [17] \textbf{Balance} pushing students to do their own work while AI is an aid. \\[3pt]
	\hdashline
	\noalign{\vskip 3pt}
	Keep human support
	& 35\newline (3.0\%)
	& Professor; TAs; in person; peers; human connection
	& [18] I think the AI learning aids were helpful, but I would have liked to have \textbf{more support from the Professor} in terms of lecturing and explaining concepts. [19] It felt very weird studying with AI instead of forming \textbf{an actual study group} and maybe \textbf{using AI as a tool}. \\[3pt]
\end{longtable}
\noindent\textit{Note.} In total, 2,019 out of 4,898 students enrolled in treatment classes responded to the survey. Among them, 1,647 responded to at least one of the open-ended questions ``What did you dislike about the AI learning aids used in this class?'' and ``How can we improve the AI learning aids?'' Responses to these two questions were combined for analyses because they overlap significantly. Among them, 1,157 were informative responses about instructors' AI design. Among the rest, 302 were noninformative responses such as non-use or positive-only responses. There were also 184 software-functionality-only responses such as ``make the log in process more smooth'' and ``allow users to save chat histories'', and 4 off-topic responses. Themes were identified through human and LLM reading of the responses with the following process. First, human reader selected a random sample of 200 responses and identified the prominent themes. LLM independently read all responses and also identified prominent themes. Human reader identified all themes listed above, whereas LLM identified the same themes but separated ``Keep Human Support'' from ``Use Less''. LLM also identified three additional detailed categories (i.e., more scaffolding, more real world problems, and answer verification and disclosure), each accounted no more than 1\% of the responses. We therefore combined them with ``clearer and deeper explanations''.  LLM then labeled all responses according to the 10 themes. To verify accuracy, the human reader labeled the 200 randomly selected responses and calculated the human-LLM classification agreement rate of 95.8\%. Theme labels were not mutually exclusive; a response could contain more than one theme. Therefore, frequencies and percentages do not sum to the sample total or 100\%. Percentages in the table are calculated as percentages of the 1,157 informative. \par
\clearpage
\endgroup

\clearpage
\begingroup
\ifdefined\nolinenumbers\nolinenumbers\fi
\setstretch{1}
\fontsize{12}{13}\selectfont
\setlength{\parskip}{0pt}
\setlength{\parindent}{0pt}
\setlength{\tabcolsep}{2pt}
\setlength{\LTleft}{0pt plus 1fill}
\setlength{\LTright}{0pt plus 1fill}
\setlength{\LTpre}{0pt}
\setlength{\LTpost}{5pt}
\setlength{\LTcapwidth}{\linewidth}
\setlength{\surveyUsableWidth}{\dimexpr\linewidth-4\tabcolsep\relax}
\setlength{\dashlinedash}{3pt}
\setlength{\dashlinegap}{2pt}
\setlength{\arrayrulewidth}{0.35pt}
\renewcommand{\arraystretch}{1}
\captionsetup[longtable]{font=normalsize,labelfont=bf,labelsep=period,
  justification=raggedright,singlelinecheck=false,skip=5pt}
\begin{longtable}{@{}
  >{\RaggedRight\arraybackslash}p{0.36\surveyUsableWidth}
  >{\Centering\arraybackslash}p{0.11\surveyUsableWidth}
  >{\RaggedRight\arraybackslash}p{0.53\surveyUsableWidth}@{}}
\caption{\textbf{Faculty intentions and reasons for using AI}}\label{tab:faculty-intentions}\\
\multicolumn{3}{@{}>{\RaggedRight\arraybackslash}p{\linewidth}@{}}{\textit{Panel A. What were you trying to do with AI, and why did you think AI would help?}}\\[5pt]
\toprule
\textbf{Theme} & \textbf{N (\%)} & \textbf{Representative keywords} \\
\midrule
\endfirsthead
\multicolumn{3}{@{}l}{\textbf{Table \thetable{} (continued)}}\\[3pt]
\toprule
\textbf{Theme} & \textbf{N (\%)} & \textbf{Representative keywords} \\
\midrule
\endhead
\midrule
\multicolumn{3}{r@{}}{\textit{Continued on next page}}\\
\endfoot
\bottomrule
\endlastfoot
\textbf{Accessible, on-demand learning support}
 & 21\newline (84\%)
 & Concept and syllabus questions; availability; accessible, low-pressure support \\*[3pt]
\midrule
\multicolumn{3}{@{}>{\RaggedRight\arraybackslash}p{\linewidth}@{}}{%
  \textbf{Professor [2]} ``I was trying to fill the gap of not being able to talk to each of them or even answer their questions about every single class. Sometimes they are embarrassed to ask things because they feel they are silly, so they ask the chatbot, because it knows all the content from the class.''
}\\[3pt]
\multicolumn{3}{@{}>{\RaggedRight\arraybackslash}p{\linewidth}@{}}{%
  \textbf{Professor [18]} ``We are trying to provide an additional resource for the students. Sometimes students may have difficulty finding a TA. Other times, they may be too shy to find a TA, or they face problems in the evening or over the weekend when a TA is not available. So this is quite a help for them.''
}\\[3pt]
\multicolumn{3}{@{}>{\RaggedRight\arraybackslash}p{\linewidth}@{}}{%
  \textbf{Professor [13]} ``I think maybe previously, if they needed to come to my office or write me emails to ask questions, and the question was not too significant, maybe they didn't want to do it because of the hassle. But with these tools available to them, the cost of asking for feedback is lower. I'm thinking that could be one way this helps them.''
}\\[3pt]
\hdashline
\noalign{\vskip 3pt}
\textbf{Scalable practice and personalized support}
 & 19\newline (76\%)
 & More practice problems; tailored explanations; personalized learning at scale \\*[3pt]
\midrule
\multicolumn{3}{@{}>{\RaggedRight\arraybackslash}p{\linewidth}@{}}{%
  \textbf{Professor [1]} ``The thing that was most exciting for me about AI was to give feedback and infinite practice problems in a way that I could endorse, that I felt would facilitate understanding and meet students where they're at. This largely took two different forms. The first was the Faculty Twin, and I set up a custom Faculty Twin instance for each of my exams. I tried to train it to follow the Socratic method and be supportive of whatever the students were able to answer, but slowly break things down and adapt to them. I think this is pretty typical for folks in our consortium, but this seemed really exciting. I could give students all the practice they wanted.''
}\\[3pt]
\multicolumn{3}{@{}>{\RaggedRight\arraybackslash}p{\linewidth}@{}}{%
  \textbf{Professor [16]} ``I think scaling for individualization is never possible without AI. There are 160 students, and how do you really personalize for everybody? Here, you have unbounded capacity, and everybody can do as much as they feel ready to do. Depending on the individual, we provide a platform, and the questions keep changing the context. I think that is the beauty of AI. I think generative AI was particularly about preparing everyone in their own way so that they touch the baseline. For me, that was a big game changer. I think the students were very happy with that.''
}\\[3pt]
\multicolumn{3}{@{}>{\RaggedRight\arraybackslash}p{\linewidth}@{}}{%
  \textbf{Professor [22]} ``As a student, what are the areas you are most interested in? What are the areas you work on the most? The AI basically goes in there and links these two. It explains blockchain, for example, if a student is a marketing student. It will link blockchain with the marketing concept at that point. [\ldots{}] So again, instead of answering the question in a generic AI model, you have to first ask: what is your background? What have you been working on?''
}\\[3pt]
\hdashline
\noalign{\vskip 3pt}
\pagebreak % keeps this theme row with its quotations (booktabs rules under arydshln allow a break after the row)
\textbf{Guided reasoning, critical thinking and AI literacy}
 & 12\newline (48\%)
 & Socratic questioning; challenging assumptions; critical thinking; AI literacy \\*[3pt]
\midrule
\multicolumn{3}{@{}>{\RaggedRight\arraybackslash}p{\linewidth}@{}}{%
  \textbf{Professor [2]} ``Those are also student-facing chatbots. We do quizzes with very simple questions. Instead of giving them a PDF with the solutions, I give them a chatbot that can discuss the solution with them. It's prompted to use the Socratic method, asking them questions instead of blurting out three paragraphs. It's the same with the midterm and final prep. For the midterm and final, we give them a set of problems that only includes the quantitative material, and then a chatbot that can guide them through it.''
}\\[3pt]
\multicolumn{3}{@{}>{\RaggedRight\arraybackslash}p{\linewidth}@{}}{%
  \textbf{Professor [3]} ``If you are talking about all the agents I build, that's the end goal. I want them to practice how they communicate and be critical. For this particular agent and its adversarial approach, the reason is that I think people think harder when they are challenged, and AI provides a safe environment for them to defend themselves.''
}\\[3pt]
\multicolumn{3}{@{}>{\RaggedRight\arraybackslash}p{\linewidth}@{}}{%
  \textbf{Professor [24]} ``Say why this course is necessary, even though AI can do everything, as one would imagine. And second, what is an instinct you should have as a manager? What things do you want to check? Where can AI go wrong? This was a pen-and-paper-based exam. It was an in-class quiz at the end of every module, a 10-minute quiz where I would give them five scenarios and the AI output for each scenario, and then ask: do you think this is perfect, or would you change something? Or what is the next question you're going to ask your AI to do?''
}\\[3pt]
\hdashline
\noalign{\vskip 3pt}
\textbf{Simulation and applied learning}
 & 10\newline (40\%)
 & Realistic interactions; simulations; case work; feedback on applied tasks \\*[3pt]
\midrule
\multicolumn{3}{@{}>{\RaggedRight\arraybackslash}p{\linewidth}@{}}{%
  \textbf{Professor [23]} ``The agent would simulate being a business owner with a problem. It would say, `Hey, I've got this problem. What do I do?' They had to interact with it to tease out the information: what is your objective, really? And what would the coefficients be? It wouldn't inform them. It would just say, `I've got this problem. I don't know what to do to make my business run better.' So the students had to tease out the information: it turns out he wants to make a product mix choice, but it's got different revenues and different cost structures. We first have to figure out how much profit is being made. They had to figure out: this is the profit we're going to be maximizing, but he doesn't give it to me. He doesn't even necessarily know what the profit is. I have to do that calculation myself.''
}\\[3pt]
\multicolumn{3}{@{}>{\RaggedRight\arraybackslash}p{\linewidth}@{}}{%
  \textbf{Professor [10]} ``I ran a simulation on day one, and that was fun. We had a leaderboard and a very gamified version of the DuPont model. I created the financial metrics of two companies, company A and company B, and at the bottom, the names of the companies, but they don't know which is which. Then they were to mix and match. I gamified that, and we had a leaderboard, so that was fun too. And in real time.''
}\\[3pt]
\multicolumn{3}{@{}>{\RaggedRight\arraybackslash}p{\linewidth}@{}}{%
  \textbf{Professor [2]} \textit{When explaining the move from a conventional written case to an AI-based case:}\newline ``They would load the PDF and ask a few questions, and that's all they did. So we thought of it as a fun way to fight fire with fire: `You're cheating this way. This is what we're going to do now.' It ended up being a lot of fun. The students liked the realistic aspect of it. We get a lot of MBA students who want to do consulting, so they said, `This really feels like consulting.' Even when they said they spent more time on the case, and it was busier than the regular cases, they liked it better.''
}\\[3pt]
\hdashline
\noalign{\vskip 3pt}
\textbf{Compatibility with existing courses and teaching approaches}
 & 6\newline (24\%)
 & Shared course structures; consistency across sections; grounding in course material \\*[3pt]
\midrule
\multicolumn{3}{@{}>{\RaggedRight\arraybackslash}p{\linewidth}@{}}{%
  \textbf{Professor [1]} ``I think this is unlike that of some others: I teach three of our 30 sections annually in core undergraduate operations management. It is a relatively standardized course. I know some other universities have even more standardization, like weekly meetings about what will be taught each week. That's not what we have, but we do have a common homework system with common homework problems. So I felt relatively constrained in rethinking everything. I didn't want to change everything about three sections out of 30. I didn't want to completely change 10\% of the sections of a class. That just doesn't seem very useful. What did I try to do? I wanted to bring AI into the classroom while keeping the structure of the course the same: the same homework and exams.''
}\\[3pt]
\multicolumn{3}{@{}>{\RaggedRight\arraybackslash}p{\linewidth}@{}}{%
  \textbf{Professor [12]} ``The big concern we had was that it would start using terms that we didn't use, or symbols that we didn't use. It would give people answers, and then they would not really learn things. Those were our biggest concerns. So we put guardrails on it---a pretty verbose, two- to three-page prompt.''
}\\[3pt]
\multicolumn{3}{@{}>{\RaggedRight\arraybackslash}p{\linewidth}@{}}{%
  \textbf{Professor [23]} ``I record all my classes, so I had all the transcripts from my prior semesters. I was able to take those transcriptions, give them to ChatGPT, have it create a summary, and upload those summaries as some of the material. I think one of the key things [\ldots{}] I saw as a strength was that it could explain the material to students in my voice, because I've got a very particular way of explaining some of this material. For example, when we're dealing with reliability calculations: how long is a satellite going to be working? You can do that formula for an exponential distribution using the time between failures or the frequency of failures. It's basically the same thing, but the formula looks slightly different. Students were sometimes using Google or AI, like ChatGPT, to look up how this works, but they would get the other formula. It's correct; it's still the same formula, but it's the inverse of what you're doing. You either put in the time between failures or the frequency of failures.''
}\\[3pt]
\hdashline
\noalign{\vskip 3pt}
\textbf{AI-informed diagnosis of student learning needs}
 & 4\newline (16\%)
 & Student conversations and attempts; conceptual gaps; instructor diagnostic insight \\*[3pt]
\midrule
\multicolumn{3}{@{}>{\RaggedRight\arraybackslash}p{\linewidth}@{}}{%
  \textbf{Professor [5]} ``Just submit that to Canvas. What was also nice is that I can actually look through them. That allows them to have some privacy: if they want to cut out some portion of the conversation, they can. So I see what they want; they can curate if needed. I can at least read them and see where people are getting tripped up. Or, even if the AI helped them, I can see that the AI was helpful in covering something, and I can generalize that. So it is really helpful to figure out what students are actually experiencing.''
}\\[3pt]
\multicolumn{3}{@{}>{\RaggedRight\arraybackslash}p{\linewidth}@{}}{%
  \textbf{Professor [24]} ``And I thought, hey, this is cool, and I would want to know what they don't understand. It was a smaller class, so I basically had the same idea implemented. Let's say there is a question with options A, B, C, D. The right answer is B, and you say A. Then I'm like, oh, that's a great way of thinking about it, but have you considered this? Try to nudge them in the direction of B rather than giving them the exact answer, and then see whether after A they went to C and then came to B, or whether after A that was enough for them to go to B, so at least I would understand where the gaps are.''
}\\[3pt]
\multicolumn{3}{@{}>{\RaggedRight\arraybackslash}p{\linewidth}@{}}{%
  \textbf{Professor [9]} ``For instance, this manufacturer is ending up with \$0 because that team set up the objective as making sure to sell the item. Exactly, same thing here: \$12,000 is so low in that simulation. They also set the objective as making sure you sell it. They don't specify any lower bound.''
}\\[3pt]
\hdashline
\noalign{\vskip 3pt}
\textbf{AI-supported preparation for classroom learning}
 & 2\newline (8\%)
 & Pre-class exercises; next-lecture previews; preparation for classroom learning \\*[3pt]
\midrule
\multicolumn{3}{@{}>{\RaggedRight\arraybackslash}p{\linewidth}@{}}{%
  \textbf{Professor [5]} ``What I ended up doing was adapting a little bit from there and bringing in some other things, but it was inspired from there. I had them do one pre-class assignment towards the end of the forecasting unit, where they are interacting virtually with Hannah Banana, who is a designer or head of something like that at a fashion goods firm. There is a whole story about that, and they are doing some more forecasting. She is finding the process really painful, so she describes what she is going through and what she set up. She describes why it is working or not working. And you are her friend: you are just meeting her for coffee, and you have to give her advice. So you are reacting and listening to what she did. For example, every year they have new clothing come out.''
}\\[3pt]
\multicolumn{3}{@{}>{\RaggedRight\arraybackslash}p{\linewidth}@{}}{%
  \textbf{Professor [21]} ``That is one of the key functions of OpsCAD, and it also previews what is going to happen in the next lecture. Again, I think in a way I could probably implement all these functions with Canvas as well. But it is awkward to navigate through Canvas. Using generative AI is more conversational, and I feel like it also has more character. So hopefully students find it useful that way.''
}\\[3pt]
\end{longtable}
\noindent\textit{Note.} These themes summarize objectives and perceived AI affordances elicited by Parts 1 and 2. N = 25 interviews. Each interview counts once per theme. Themes may overlap. Professor IDs refer to the revised participant register. Quotations have been lightly edited for clarity and readability, including removal of fillers, repetitions and false starts, without changing the intended meaning. [\ldots{}] indicates omitted passages. Adjacent turns by the same speaker are joined.\par

\clearpage

\begin{longtable}{@{}
  >{\RaggedRight\arraybackslash}p{0.36\surveyUsableWidth}
  >{\Centering\arraybackslash}p{0.11\surveyUsableWidth}
  >{\RaggedRight\arraybackslash}p{0.53\surveyUsableWidth}@{}}
\caption{\textbf{Faculty experiences during AI implementation}}\label{tab:faculty-implementation}\\
\multicolumn{3}{@{}>{\RaggedRight\arraybackslash}p{\linewidth}@{}}{\textit{Panel B. What actually happened when you implemented it? How did students actually use or interact with the AI? Did they use it in the way you expected? Was there anything that worked differently than expected, surprised you, or caused you to change the intervention during the semester?}}\\[5pt]
\toprule
\textbf{Theme} & \textbf{N (\%)} & \textbf{Representative keywords} \\
\midrule
\endfirsthead
\multicolumn{3}{@{}l}{\textbf{Table \thetable{} (continued)}}\\[3pt]
\toprule
\textbf{Theme} & \textbf{N (\%)} & \textbf{Representative keywords} \\
\midrule
\endhead
\midrule
\multicolumn{3}{r@{}}{\textit{Continued on next page}}\\
\endfoot
\bottomrule
\endlastfoot
\textbf{Useful AI support in practice}
 & 19\newline (76\%)
 & Useful explanations and task help; additional and varied practice; positive reception \\*[3pt]
\midrule
\multicolumn{3}{@{}>{\RaggedRight\arraybackslash}p{\linewidth}@{}}{%
  \textbf{Professor [1]} ``In the expected way, I knew what students wanted: they wanted more practice. Providing tools that offered more practice, I think, made them better prepared, or at the very least made them not complain to me about it. [\ldots{}] It succeeded on that front: it let students who wanted to prepare more do so, and those who didn't just had to live with that decision.''
}\\[3pt]
\multicolumn{3}{@{}>{\RaggedRight\arraybackslash}p{\linewidth}@{}}{%
  \textbf{Professor [11]} ``Not all students engaged with the AI I trained. However, many of those who engaged with it found it useful. They especially used the apps to get numerous examples from different contexts. Providing examples for the students to practice with was very time-consuming before using AI. The examples were limited, and contexts were rather standard. AI made these diverse and made it possible to scale them up. There were no big surprises.''
}\\[3pt]
\multicolumn{3}{@{}>{\RaggedRight\arraybackslash}p{\linewidth}@{}}{%
  \textbf{Professor [15]} ``And because the way we do our notes and our exams, at least back then, were derivatives of what we had done in class, I think the AI was pretty good at handling that. So you gave it a new question that was very similar to something we had done in the notes or in the homework, and then the AI could adapt to that. It was exactly what the AI was meant to do: adapt things to a particular context with slight changes in numbers. So I think it worked really well for them to ask it to explain certain things.''
}\\[3pt]
\hdashline
\noalign{\vskip 3pt}
\textbf{Uneven and context-dependent AI use}
 & 13\newline (52\%)
 & Uneven use; lower-than-expected uptake; optional versus required participation \\*[3pt]
\midrule
\multicolumn{3}{@{}>{\RaggedRight\arraybackslash}p{\linewidth}@{}}{%
  \textbf{Professor [20]} ``The other thing I noticed is that I think it was very skewed: some people would use it a lot. For the rest, it was a one-time thing, and that was it.''
}\\[3pt]
\multicolumn{3}{@{}>{\RaggedRight\arraybackslash}p{\linewidth}@{}}{%
  \textbf{Professor [3]} ``I learned that throughout the semester. At the beginning, I thought these tools were helpful, so I put them out there for students to use voluntarily. But I found that students don't use them unless you give bonus points or tie them to a particular class requirement. So, for the two agents, I do give points, and I feel that it is a must for students of this age. If you don't require them to use it, even though it's beneficial, I think they're optimizing their grades using the least amount of time. So probably not many people will use it.''
}\\[3pt]
\multicolumn{3}{@{}>{\RaggedRight\arraybackslash}p{\linewidth}@{}}{%
  \textbf{Professor [14]} ``The actual usage is much lower than I expected. [\ldots{}] For the weekly assignment in particular, students just use their own AI tools. Their own AI tools are already good enough; the AI agents do not have to be trained on that week's content to help them with the weekly assignments. That was beyond my expectation. But other than that, I think the students are using the practice exam AI very extensively.''
}\\[3pt]
\hdashline
\noalign{\vskip 3pt}
\textbf{Technical friction and ongoing refinement}
 & 9\newline (36\%)
 & Prompt adjustments; technical limitations; refining how the tutor responds \\*[3pt]
\midrule
\multicolumn{3}{@{}>{\RaggedRight\arraybackslash}p{\linewidth}@{}}{%
  \textbf{Professor [25]} ``I changed the prompt several times, but it kept giving out details and answering discussion questions. So I ended up not using the case [\ldots{}] chatbot.''
}\\[3pt]
\multicolumn{3}{@{}>{\RaggedRight\arraybackslash}p{\linewidth}@{}}{%
  \textbf{Professor [21]} ``Summarizing what happens in today's class: this is where I think the AI agent is very useful. Usually, right after I finish class, I update [\ldots{}] the agent by changing the prompts, so I can customize it and try to clarify things---maybe I wasn't being very clear in class. [\ldots{}] Or maybe I observe students asking me about certain points after class. I try to clarify those points using the agent, by highlighting them and providing short explanations. I write and update these myself, manually.''
}\\[3pt]
\multicolumn{3}{@{}>{\RaggedRight\arraybackslash}p{\linewidth}@{}}{%
  \textbf{Professor [9]} ``In about week five, I had to intervene and add a prompt to make sure it would never release the answers to any quiz material or other questions from the class. After that [\ldots{}] usage has come down a little.''
}\\[3pt]
\hdashline
\noalign{\vskip 3pt}
\textbf{Changes in human support use}
 & 6\newline (24\%)
 & Office hour attendance; email frequency; instructor-student interaction \\*[3pt]
\midrule
\multicolumn{3}{@{}>{\RaggedRight\arraybackslash}p{\linewidth}@{}}{%
  \textbf{Professor [6]} ``As I said, it was trained on course, the syllabus and materials, so students could ask questions about the course and logistics, 24/7. In my experience, I feel there has been a significant reduction in the number of students emailing me about course logistics or coming to my office hours.''
}\\[3pt]
\multicolumn{3}{@{}>{\RaggedRight\arraybackslash}p{\linewidth}@{}}{%
  \textbf{Professor [7]} ``I'm not sure whether this is a positive or negative thing from a time-use perspective. This is perhaps a positive or more efficient signal, but it also reduces interaction between instructors and students. In some way, they all turn to AI.''
}\\[3pt]
\multicolumn{3}{@{}>{\RaggedRight\arraybackslash}p{\linewidth}@{}}{%
  \textbf{Professor [14]} ``I don't think so. Typically, for elective courses, people don't have that many questions to bring to office hours. At least from what I heard from the TAs, I do not see an increase or decrease compared with previous years.''
}\\[3pt]
\hdashline
\noalign{\vskip 3pt}
\textbf{Task-oriented AI use with limited exploration}
 & 3\newline (12\%)
 & Quiz-focused use; activity around deadlines; limited exploratory engagement \\*[3pt]
\midrule
\multicolumn{3}{@{}>{\RaggedRight\arraybackslash}p{\linewidth}@{}}{%
  \textbf{Professor [9]} ``They only used it to solve the quizzes. My objective was to have something they could talk to, engage with and ask [\ldots{}] more sophisticated questions, and learn about something. That wasn't the case for the majority of them.''
}\\[3pt]
\multicolumn{3}{@{}>{\RaggedRight\arraybackslash}p{\linewidth}@{}}{%
  \textbf{Professor [22]} ``The project agents are a huge hit. They have very high use, which tends to increase before exams. As assignment deadlines approach, students use them more and more.''
}\\[3pt]
\multicolumn{3}{@{}>{\RaggedRight\arraybackslash}p{\linewidth}@{}}{%
  \textbf{Professor [24]} ``That was, I think, very surprising to me, given that the conversations in class and around them must have been around AI. I saw big spikes around assignments, and that was a trend I continued to see across these two different implementations of AI over the two years of this experiment.''
}\\[3pt]
\end{longtable}
\noindent\textit{Note.} Part 3 prompted accounts of student use, expectations, surprises and adjustments, not a comprehensive assessment of whether intended benefits were achieved. Human support accounts include reports of change and no change. N = 25 interviews. Each interview counts once per theme. Themes may overlap. Professor IDs refer to the revised participant register. Quotations have been lightly edited for clarity and readability, including removal of fillers, repetitions and false starts, without changing the intended meaning. [\ldots{}] indicates omitted passages. Adjacent turns by the same speaker are joined.\par

\clearpage
\endgroup

\end{document}